\documentclass[sigconf]{acmart}
\AtBeginDocument{%
  }

\setcopyright{acmlicensed}
\copyrightyear{2026}
\acmYear{2026}
\acmDOI{XXXXXXX.XXXXXXX}
\acmConference[KDD '26]{Make sure to enter the correct
  conference title from your rights confirmation email}{June 03--05,
  2026}{Woodstock, NY}
\acmISBN{978-1-4503-XXXX-X/2018/06}

\usepackage{balance}
\usepackage{amsmath}

\usepackage{amssymb}
\usepackage{mathtools}
\usepackage{amsthm}
\usepackage[noend,algo2e,noline,ruled,linesnumbered]{algorithm2e}

\usepackage[capitalize,noabbrev]{cleveref}

\usepackage{subcaption}
\usepackage{url}

\theoremstyle{plain}
\newtheorem{theorem}{Theorem}[section]
\newtheorem{proposition}[theorem]{Proposition}

\theoremstyle{definition}

\theoremstyle{remark}

\usepackage[textsize=tiny]{todonotes}

\usepackage[textsize=tiny]{todonotes}

\newcommand{\mbx}[0]{\mathbf{x}}
\newcommand{\mbz}[0]{\mathbf{z}}
\newcommand{\mbX}[0]{\mathbf{X}}
\newcommand{\mcY}[0]{\mathcal{Y}}
\newcommand{\mcX}[0]{\mathcal{X}}

\newcommand{\mcD}[0]{\mathcal{D}}
\newcommand{\BT}[0]{\texttt{BT}}
\newcommand{\TM}[0]{\texttt{TM}}

\newcommand{\selrisk}[0]{R_{l}(f,g)}

\DeclareMathOperator*{\argmax}{arg\,max}
\DeclareMathOperator*{\argmin}{arg\,min}
\newcommand{\riskLowerBeta}[0]{r(\mcX\times\mcX)_{<\beta}}
\newcommand{\riskEqualBeta}[0]{{r(\mcX\times\mcX')_{=\beta}}}
\newcommand{\riskLowerA}[0]{r(\mcX\times\mcX)_{<a}}
\newcommand{\riskEqualA}[0]{{r(\mcX,\mcX')_{=a}}}
\newcommand{\riskGreaterBeta}[0]{r(\mcX\times\mcX')_{>\beta}}
\newcommand{\riskGreaterA}[0]{r(\mcX,\mcX')_{>a}}
\newcommand{\riskGreaterEqualBeta}[0]{r(\mcX,\mcX')_{\geq\beta}}

\newcommand{\BALToR}{\texttt{BALToR}}
\newcommand{\BALToRc}{\texttt{BALToR-conf}}
\newcommand{\BALToRm}{\texttt{BALToR-marg}}
\newcommand{\BALToRe}{\texttt{BALToR-entr}}

\newcommand{\pyxx}[0]{p(y\mid\mbx, \mbx')}

\title{Bounded-Abstention Pairwise Learning to Rank}

\author{Antonio Ferrara}
\authornote{Both authors contributed equally to this work.}
\email{antonio.ferrara@intesasanpaolo.com}
\orcid{0000-0002-0784-1115}
\affiliation{%
  \institution{Intesa Sanpaolo AI Research}
  \city{Turin}
  \country{Italy}
}

\author{Andrea Pugnana}
\authornotemark[1]
\email{andrea.pugnana@unitn.it}
\orcid{0000-0001-9138-8212}
\affiliation{%
  \institution{University of Trento}
  \city{Trento}
  \country{Italy}
}

\author{Francesco Bonchi}
\email{francesco.bonchi@intesasanpaolo.com}
\orcid{0000-0001-9464-8315}
\affiliation{%
  \institution{Intesa Sanpaolo AI Research}
  \city{Turin}
  \country{Italy}
}

\author{Salvatore Ruggieri}
\email{salvatore.ruggieri@unipi.it}
\orcid{0000-0002-1917-6087}
\affiliation{%
  \institution{University of Pisa}
  \city{Pisa}
  \country{Italy}
}

\copyrightyear{2026}
\acmYear{2026}
\setcopyright{cc}
\setcctype{by}
\acmConference[KDD '26]{Proceedings of the 32nd ACM SIGKDD Conference on Knowledge Discovery and Data Mining V.2}{August 09--13, 2026}{Jeju Island, Republic of Korea}
\acmBooktitle{Proceedings of the 32nd ACM SIGKDD Conference on Knowledge Discovery and Data Mining V.2 (KDD '26), August 09--13, 2026, Jeju Island, Republic of Korea}
\acmDOI{10.1145/3770855.3817918}
\acmISBN{979-8-4007-2259-2/2026/08}

\begin{document}



\begin{abstract}
Ranking systems influence decision-making in high-stakes domains like health, education, and employment, where they can have substantial economic and social impacts. This makes the integration of safety mechanisms essential. One such mechanism is \textit{abstention}, which enables algorithmic decision-making systems to defer uncertain or low-confidence decisions to human experts.
While abstention has been predominantly explored in the context of classification tasks, its application to other machine learning paradigms remains underexplored. In this paper, we introduce a novel method for abstention in pairwise learning-to-rank tasks. Our approach is based on thresholding the ranker's conditional risk: the system abstains from making a decision when the estimated risk exceeds a predefined threshold. Our contributions are threefold: a theoretical characterization of the optimal abstention strategy, a model-agnostic, plug-in algorithm for constructing abstaining ranking models, and  a comprehensive empirical evaluation across multiple datasets, demonstrating the effectiveness of our approach.

\end{abstract}




\maketitle
\sloppy 

\section{Introduction}
\begin{figure*}[t!]
    \centering
    \includegraphics[width=.9\linewidth]{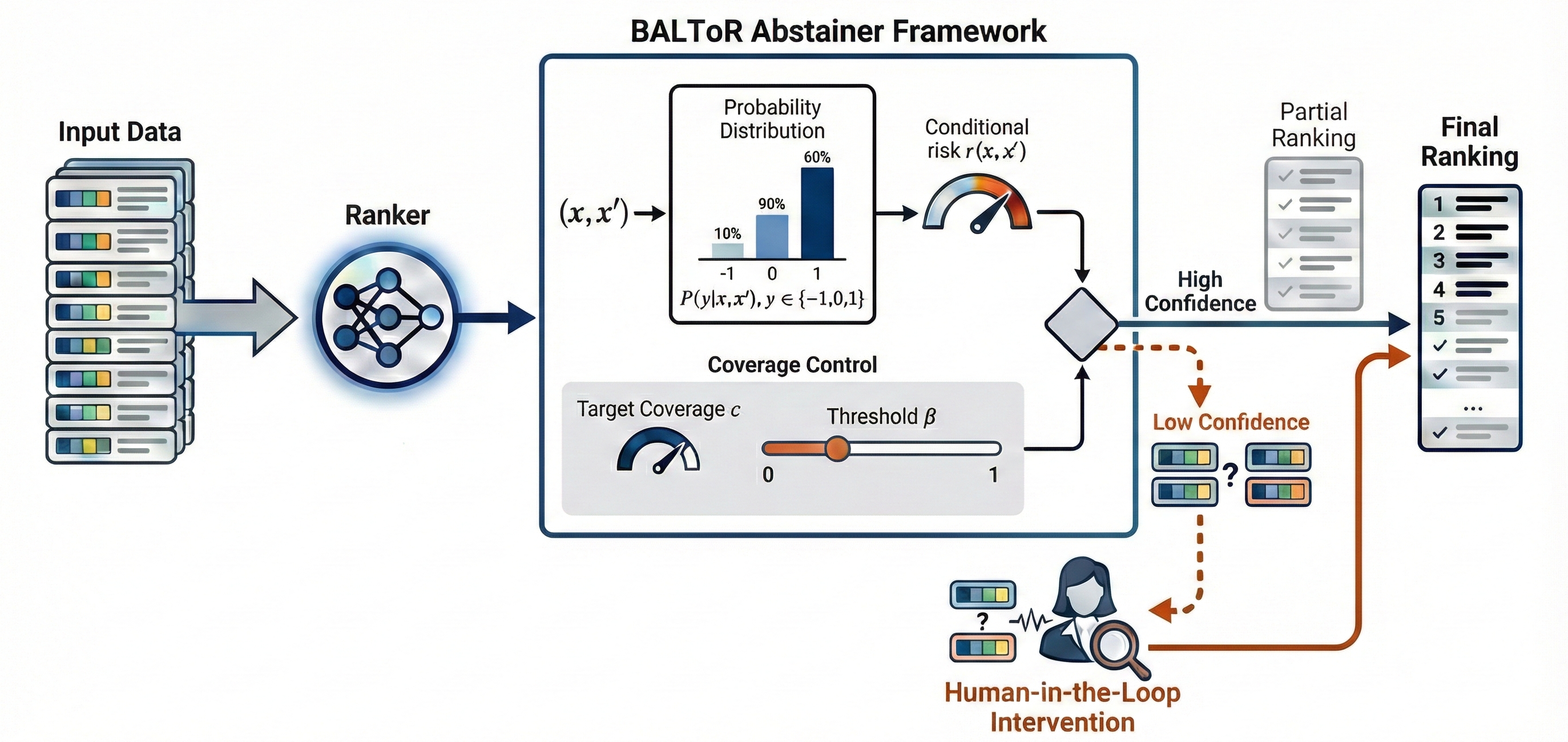}
    \vspace{-3mm}
    \caption{Illustration of the role of the \BALToR\ abstainer in a ranking process. A ranker produces a complete ordering of items for a given query. However, the ranker may exhibit low confidence in the relative ordering of certain item pairs. \BALToR\ identifies a number of these low-confidence pairs and defer them to an evaluator for further inspection. Such an evaluator - a human or even another model - leverages additional knowledge or higher-quality data to resolve the uncertainty on those pairs and, finally, producing a more reliable final ranking.}
    \label{fig:example}
\end{figure*}
Ranking systems have become deeply embedded in modern decision-making processes, affecting outcomes in 
%
high-stakes domains such as healthcare (e.g., prioritizing patients for clinical trials or organ transplants), education (e.g., university admission), and employment (e.g., job candidate screening). 
Because an individual's outcome is often determined by their position in the ranking, these algorithmic decision-making systems can exert profound economic and social influence.
As a result, it is essential to equip them with robust safety mechanisms.
Among the emerging forms of \emph{``AI guardrails'}', abstention mechanisms 
enable AI systems to refrain from issuing a low-confidence prediction.
This allows for either direct human intervention, such as making a judgment on a specific case, or for the collection of additional, higher-quality data to be fed as input to the system.

So far, abstention, also referred to as ``rejection", has been primarily studied in the context of classification tasks (see \S \ref{sec:background_abs} for the relevant literature), while receiving limited attention in other machine learning (ML) settings.
In this paper, we address the problem of \emph{bounded abstention in pairwise learning to rank with ties}. The objective of pairwise learning to rank is to train a model capable of predicting pairwise preferences between item pairs for a given query. For instance, in a hiring scenario, the query is a job opening, and the two items to be compared are two candidates. Notice that building a classifier can easily become computationally unfeasible when considering all the pairs of instances. Hence, one needs to resort to learning-to-rank approaches, which derive a global ranking from a dataset of pairwise preferences (see \S \ref{subsec:background1}). Therefore, given a trained ranker, i.e., a ML model trained for ranking, our goal is to build ad-hoc abstention strategies.

Regarding the abstention strategies, two main approaches are generally considered in the literature~\citep{DBLP:journals/jmlr/FrancPV23}: assigning a fixed cost to each abstention in the loss function (\emph{cost-based model}) or balancing the fraction of allowed abstentions against the improvement in the ML model's performance (\emph{bounded-abstention model}).
\citet{Mao23} provide a theoretical analysis of the cost-based formulation in a pairwise ranking setting. However, in practice, establishing the cost of abstaining compared to making an error is very challenging. 
We focus on the bounded-abstention formulation, where a fixed fraction of abstentions is specified in advance, and the goal is to optimize performance within such a constraint.
This approach enables resource management, especially in scenarios where human decision-making capacity is limited. E.g., in a recruitment setting, a company may allocate a specific budget for additional data collection or expert review to handle uncertain predictions made by the ML system. This budget translates into a predefined threshold on the number of abstentions the system is allowed to make.

Figure \ref{fig:example} illustrates our setting. 
Given a set of items (each represented by a feature vector) to be ranked for a specific query, a standard ranker would output a complete ordering of the items. In contrast, our approach introduces an \textit{abstainer} that identifies item pairs for which the ranker’s confidence is low. These uncertain comparisons are then (possibly) deferred to a human expert for further evaluation. 
We propose \BALToR{} (\textbf{B}ounded-\textbf{A}bstention \textbf{L}earning \textbf{To} \textbf{R}ank), a novel method for abstention in pairwise learning-to-rank settings. To devise \BALToR{}, we first characterize the optimal selection strategy that determines when the ranking model should abstain from providing a relative ordering between item pairs. 
The strategy is based on thresholding the model’s conditional risk: the system abstains when the estimated risk exceeds a predefined threshold. Given a bound on the allowable fraction of abstentions over pairwise comparisons, \BALToR{} aims to maximize ranking performance on the remaining, non-abstained pairs.
%

Our contributions can be summarized as follows:
\begin{itemize}
    \item [$(i)$] We formally introduce the problem of \textit{bounded-abstention pairwise learning-to-rank} and provide a theoretical characterization of the optimal abstention strategy when \textit{the ranker is given}, a common setting in the real world when retraining the ranker is computationally too expensive (e.g., for foundation models);
    \item [$(ii)$]  We propose a \textit{model-agnostic} plug-in algorithm for building an abstaining ranking model. Our approach relies on the optimal strategy of $(i)$, and determines which pairs of instances the ranker should abstain on; 
    \item [$(iii)$] We conduct an empirical evaluation of our approach across multiple datasets and alternative heuristics, showing the effectiveness of the proposed methodologies.
\end{itemize}

The paper is structured as follows. In \S \ref{sec:background}, we introduce the frameworks of  pairwise learning to rank and  bounded abstention, highlighting the distinction of our work with the existing literature. In \S \ref{sec:method}, we describe the proposed  method for bounded-abstention pairwise learning to rank, present the theoretical characterization of the optimal abstention strategy, and provide the algorithmic details. In \S \ref{sec:experiments}, we empirically evaluate the proposed method on various datasets. Finally, in \S \ref{sec:conclusion}, we conclude and outline limitations and future directions.

\section{Background and Related Work}
\label{sec:background}
Let $\mcX$ be an \textit{input space}, $\mcY$ the \textit{target space} and $P$ a (unknown) joint probability distribution over $\mcX\times \mcY$. Given a \textit{hypothesis space} $\mathcal{F}$ of functions that map $\mcX$ to $\mcY$, the goal of a ML algorithm (a \textit{learner}) is to find a hypothesis $f \in\mathcal{F}$ that minimizes the \textit{risk}: \[ R(f) = 
\mathbb{E}_{\mbX, Y \sim P}[l(f(\mbX), Y)] \]
where  $l:\mcY \times \mcY \to \mathbb{R}_{\geq 0}$ is a user-specified \textit{loss function}. 

\subsection{Pairwise Learning to Rank with Ties}
\label{subsec:background1}

We consider the standard pairwise learning to rank scenario where a set of queries \( Q = \{q_1, q_2, \ldots, q_n\} \) and, for each query \( q \), a set of item-query couples \( D^q = \{d^q_1, d^q_2, \ldots, d^q_m\} \) are given. 
Typically, for each item-query couple, one has access to a set of features $\mbx^q\in\mcX$, representing characteristics of the item for that query. 
For instance, in a hiring scenario, a query is a job offer, and items are candidates. Features may regard attributes on a candidate (e.g., age) or also on the matching of the candidate to the job offer (e.g., share of job offer skills fulfilled by the candidate).
Furthermore, given a pair of items for the same query, we call it a pair of instances $\mbx^q,\mbx'^q\in \mcX\times\mcX$, and we define pairwise preference as an ordering relationship ($\succ_q, \prec_q, =_q$) where \( d^q_{\mbx} \succ_q d^q_{\mbx'} \) indicates that \( d^q_{\mbx} \) is preferred over \( d^q_{\mbx'} \) for query \( q \), \( d^q_{\mbx} \prec_q d^q_{\mbx'} \) that \( d^q_{\mbx'} \) is preferred over \( d^q_{\mbx} \), and $=_q$ indicates a tie. 
We consider a set of pairwise preferences given for each training query. Alternatively, a relevance score might be given for each document-query couple, from which pairwise preferences among two items for the same query are inferred.
This preference ordering can be encoded by considering a target space $\mcY=\{-1,0,1\}$, where $-1$ means that the second instance is preferred (or ranked higher) than the first, $1$ indicates that the first is preferred and $0$ a tie. From now on, for easing the notations, we will omit the query indicators.

The objective of pairwise learning to rank is to learn a ranker $f:\mcX\times\mcX\to \mcY$, which models the relative rankings between all item pairs for each query. 
The canonical loss one aims to minimize in such a setting is the mis-ranking loss, which is defined as $l_r(f(\mbx,\mbx'), y) = \mathbb{I}_{y\neq f}$.
In other words, $f$ incurs a loss of one whenever the pair $(\mbx,\mbx')$ is ranked by  $f$ differently from the ground truth order; if the ranking is correct, no loss is incurred.
Let $P$ be a probability distribution over $\mcX \times \mcX \times \mcY$. We denote with $R_{l}(f) = \mathbb{E}_{(\mbX,\mbX', Y)\sim P}[l\left(f (\mbX, \mbX'), Y\right)]$
the expected $l$ loss of the ranker $f$. Typically, such a loss is not directly tractable. Hence, a few approaches propose to learn scores $s:\mcX\to\mathbb{R}$ that provide a value for the relevance of the single item $\mbx$. Generally, these scores are learnt through (stochastic) gradient descent and boosting techniques, e.g., Ranknet, LambdaRank and LambdaMart \citep{burges2010ranknet}.

\subsection{Abstaining Machine Learning}
\label{sec:background_abs}
Abstention mechanisms have been predominantly explored in the context of classification \citep{DBLP:conf/nips/GeifmanE17,DBLP:conf/icml/GeifmanE19,DBLP:conf/uai/SchreuderC21,DBLP:conf/aaai/PugnanaR23,DBLP:conf/aistats/PugnanaR23,DBLP:conf/iclr/FengAHA23,DBLP:conf/pkdd/LendersPPCPG24,rabanser25}, while limited attention has been devoted to other ML tasks~\citep{DBLP:conf/nips/PeriniD23,DBLP:conf/ecai/StradiottiPD24,stradiotti26}.

The risk of a predictor can be controlled through a selection function $g:\mcX\to\{0,1\}$ that determines whether the predictor shall be invoked ($g(\mathbf{x})=1$) or it should abstain from producing an output ($g(\mathbf{x})=0$) \citep{DBLP:journals/amai/CortesDM24}. The former case is also denoted as accepting an instance, and the latter as rejecting the instance.
Abstention mechanisms include \textit{novelty rejection}~\citep{DBLP:conf/iciap/CordellaSSV95}, which focuses on rejecting instances out of the training distribution, and \textit{ambiguity rejection}~\citep{DBLP:journals/ml/HendrickxPPMD24}, which focuses on rejecting instances close to the decision boundary of the predictor. 
For approaches combining both strategies, see \citet{pruuvsaconstrained}.
This paper belongs to the ambiguity rejection research line, for which two main frameworks have been considered. 
Learning to Reject~\citep{DBLP:journals/tit/Chow70} considers a loss function parametric in the cost $a$ of rejecting an instance~\citep{Herbei06,DBLP:journals/prl/Tortorella05,DBLP:conf/isbi/CondessaBCOK13,DBLP:conf/nips/CortesDM16}.
However, determining the value of $a$ is context-dependent and difficult to determine \citep{denis2020consistency,DBLP:conf/aaai/RuggieriP25}. Moreover, there is no \textit{ex-ante} control on the fraction of abstentions, which is a critical requirement in practice. 
Selective Prediction methods solve those concerns by controlling the trade-off between performance and abstention through the \textit{coverage} $\phi(g) = \mathbb{E}[g(\mbX)]$~\citep{DBLP:journals/jmlr/El-YanivW10}, namely the probability of acceptance. 
In the bounded-abstention formulation~\citep{DBLP:journals/jmlr/FrancPV23}, this translates into the following objective:
\begin{equation}
    \argmin_{(f,g)} 
    \frac{\mathbb{E}_{\mbX,Y\sim P}[l\left(f(\mbX),Y \right)g(\mbX)]}{\phi(g)} \ \ \text{s.t.} \ \ \phi(g)\geq c,
\end{equation}
where $c$ is the \textit{target coverage}, i.e., the user-defined fraction of cases for which the selective predictor $(f, g)$ \textit{must} provide a prediction. Theoretical results for the regression task can be found in \citet{DBLP:conf/nips/ZaouiDH20}. Our Theorem \ref{thm:opt} extends the result for probabilistic classifiers by \citet{DBLP:journals/jmlr/FrancPV23} to the pairwise-learning-to-rank-with-ties task.

To the best of our knowledge, in the setting of ranking, only the following approaches have been proposed: \citet{cheng2010predicting,DBLP:conf/nips/ChengHWW12} study the problem of abstention, in particular for the label ranking setting, with the selective predictors returning partial orders (instead of total orders) on the labels;  \citet{Mao23} have considered pairwise ranking with abstention in the framework of Learning to Reject, introducing a cost for abstaining on a pair of instances. 
As for classification, such an approach hinders a practical implementation because specifying the correct cost for abstaining on a pair is not straightforward, making the cost-based model and the bounded-abstention approaches are not directly comparable~\citep{Benchmark2024}.
Our work complements theirs by: $(i)$ formulating selective pairwise ranking as a bounded-abstention problem, with advantages in terms of practical usage; $(ii)$ extending the setting to pairwise ranking with ties, providing optimal strategies also in that context.

\subsection{Differences with Classification}

Furthermore, we highlight several differences and properties that distinguish our work from abstention mechanisms on classification.

\noindent
\textbf{Scalability}: Training a classifier over pairs of instances and their permutations is unfeasible in many cases, especially on the learning to rank datasets which are often large-scale datasets.

\noindent
\textbf{Antisymmetry}: One needs to explicitly enforce that $f(x,x')=-f(x',x)$, which is not generally handled by standard classifiers.

\noindent
\textbf{Transitivity}: In ranking settings (even with ties) one requires the ranker to be transitive across multiple items, i.e., if item $x$ is better than $x'$  and  is better than $x''$, then $x$ must be better than $x''$. Resorting to a classification strategy does not guarantee that learning a full ranking (with ties) over pairs is possible. 

\noindent
\textbf{Applicability to ranking with general losses}: our approach takes as input an already trained ranker, which could also be trained with any loss (e.g., NDCG); hence, we are solving a distinct problem compared to standard classification-based abstention mechanisms.

\section{Bounded Abstention Learning To Rank}
\label{sec:method}
We adopt a \textit{selection function} $g:\mcX\times\mcX\to\{0,1\}$ to decide whether a ranker $f$ should abstain (value $0$) or should output (value $1$) its prediction on the relative order of two input instances.
A \textit{selective ranker} is a pair $(f, g)$, that abstains if $g(\mbx, \mbx')=0$, and returns $f(\mbx, \mbx')$ if $g(\mbx, \mbx')=1$.
We extend the bounded-abstention paradigm of classification (see \S\ref{sec:background_abs}), by defining the (pairwise) coverage $\phi(g) = \mathbb{E}[g(\mbX,\mbX')]$ as the expected fraction of \textit{pairs of instances} for which the ranker provides a prediction.
We define the selective risk as follows:
\[ \selrisk = \frac{\mathbb{E}_{\mbX, \mbX', Y \sim P}\left[l\left((f(\mbX, \mbX'), Y\right)g(\mbX,\mbX')\right]}{\phi(g)}\]
where $P$ is a (unknown) distribution over $\mcX\times\mcX\times \mcY$.
For a given ranker $f$, the bounded-abstention pairwise learning to rank problem is to find a selection function from a space of hypotheses $\mathcal{G}$ that minimizes selective risk while achieving a (minimum) target coverage $c$:
\begin{equation}
\label{eq:l2rank_abst}
\argmin_{g\in\mathcal{G}}\selrisk\quad \text{s.t.} \quad \phi(g)\geq c
\end{equation}

Let us now characterize a solution to the problem (\ref{eq:l2rank_abst}). We write $p(\mbx, \mbx', y)$ for the density of $P$.
We denote the conditional expected risk on the pair $(\mbx,\mbx')$ by:
\[ r(\mbx, \mbx') = \sum_{y\in{Y}}l\left(f(\mbx, \mbx'),y\right) p(y\mid \mbx, \mbx') \]
Moreover, 
we define the subspaces of $\mcX\times \mcX$ where the conditional risk is lower, equal, or greater than $a\in\mathbb{R}$ respectively as~$\riskLowerA= \{(\mbx, \mbx') \in \mcX\times\mcX: r(\mbx, \mbx') < a\}$;~$\riskEqualA = \{(\mbx, \mbx') \in \mcX\times\mcX: r(\mbx, \mbx') = a\}$ and
$\riskGreaterA = \{(\mbx, \mbx') \in \mcX\times\mcX: r(\mbx, \mbx') >a\}$.
We assume that the loss is \textit{symmetric}, namely~$l\left(f(\mbx,\mbx'), y\right) = l\left(f(\mbx', \mbx), -y\right)$.
This is a natural assumption in most contexts, since stating that $\mbx$ is better than $\mbx'$ ($d^q_{\mbx} \succ_q d^q_{\mbx'}$) is equivalent to state that $\mbx'$ is worse than $\mbx$ ($d^q_{\mbx'} \prec_q d^q_{\mbx}$), and, \textit{a fortiori}, mistakes in predicting the relative rankings are symmetric.
We can characterize now the solutions to problem (\ref{eq:l2rank_abst}).
\begin{theorem}
\label{thm:opt}
Let us consider a symmetric loss.
    A selection function $g^*:\mcX\times \mcX\to [0,1]$ is an optimal solution to Eq.~\ref{eq:l2rank_abst} if and only if the following conditions hold:
    \begin{equation}
            \begin{split}
        \iint_{\riskLowerBeta} p(\mbx,\mbx')g^*(\mbx,\mbx')d\mbx d\mbx' =\\\iint_{\riskLowerBeta} p(\mbx,\mbx')d\mbx d\mbx';
        \end{split}
        \tag{p1}
        \end{equation}
\begin{equation}\begin{split}\iint_{\riskEqualBeta} p(\mbx,\mbx')g^*(\mbx,\mbx')d\mbx d\mbx' =\\ c-\iint_{\riskLowerBeta} p(\mbx,\mbx')d\mbx d\mbx';
        \end{split}
        \tag{p2}
        \end{equation}
        \begin{equation}
\begin{split}\iint_{\riskGreaterBeta} p(\mbx,\mbx')g^*(\mbx,\mbx')d\mbx d\mbx' =0,
        \end{split}
        \tag{p3}
    \end{equation}
where $\beta = \inf \left\{a: 
 \iint_{\riskLowerA} p(\mbx,\mbx')d\mbx d\mbx' \geq c \right\}$.
\end{theorem}
\begin{proof}
    The proof is provided in \cref{prf:thm_3_1}.
\end{proof}
Intuitively, $\beta$ is the $c$-th conditional-risk quantile.
The conditions of the theorem can be read as: we should abstain on those pairs where the conditional risk is above $\beta$ (p3), and accept those where it is below $\beta$ (p1). For pairs with conditional risk equal to $\beta$, the abstention is random with a probability that satisfies (p2). 
Moreover, notice that assuming a symmetric loss guarantees that we reject consistently both $(\mbx, \mbx')$ and $(\mbx', \mbx)$ pairs for a given $\beta$.

This intuition is further formalized in the following result:

\begin{theorem}
\label{thm:optimal_solution}
Let us consider a symmetric loss.
 Consider the selection function:
 \begin{equation}
     g^*(\mbx,\mbx')=
     \begin{cases}
         1 \quad \text{if}\quad r(\mbx,\mbx')<\beta\\
         \xi\quad \text{if}\quad r(\mbx,\mbx')=\beta\\
         0\quad \text{if}\quad r(\mbx,\mbx')>\beta,
     \end{cases}
 \end{equation}
 where $\xi = 0$ if
$$ \iint_{\riskEqualBeta}p(\mbx,\mbx')d\mbx d\mbx' =0$$ and $\xi = \texttt{Bernoulli}(p_r)$ otherwise, 
where the probability $p_r$ is defined as: $$p_r = \frac{c-\iint_{\riskLowerBeta}p(\mbx,\mbx')d\mbx d\mbx'}{\iint_{\riskLowerBeta}p(\mbx,\mbx')d\mbx d\mbx'}$$
Then,  $g^*$ satisfies conditions (p1, p2, p3) of Theorem \ref{thm:opt}, and, thus, it is an optimal solution of Eq. \ref{eq:l2rank_abst}.
\end{theorem}
\begin{proof}
    The proof is provided in~\cref{prf:thm_3_2}.
\end{proof}

Theorem \ref{thm:optimal_solution} states that implementing an optimal rejection strategy requires thresholding the true conditional risk. However, such a quantity must be estimated in practice. Hence, we propose different measures to approximate that. First, notice that in the case of the 0-1 loss, we have the following closed form of conditional risk for the Bayes optimal ranker:
\begin{proposition}
\label{thm:0-1lossSpecial}
Let us define the Bayes optimal ranker as: $f^*(\mbx, \mbx') = \argmax_{y\in\mcY} \pyxx$.
Its conditional risk w.r.t. the 0-1 loss is:
    \begin{equation}
        r_{0-1}(\mbx,\mbx') = 1-\max_{y\in\mcY}\pyxx
    \end{equation}
\end{proposition}
\begin{proof}
    The proof is provided in~\cref{prf:pro_3_3}.
\end{proof}

Hence, a first natural way to estimate the conditional risk is by applying a standard plug-in approach to the quantities above, i.e., define $\hat{p}_y$ as the estimated predicted probability for $y$ and use $\max_{y\in\mcY}\hat{p}_y$ (the \textit{confidence}) as the conditional risk estimate. 
Clearly, if the model is properly calibrated, i.e., $\hat{p}_{max} \approx p_{ymax}$, using the confidence directly estimates the conditional risk. However,  other heuristics can be considered: one can resort to the estimated \textit{margin}\footnote{Notice that the ranking over instances induced by margin and the confidence can differ in the pairwise ranking with ties, as it is immediate to show that $1-margin \geq 1 - p_{max}$.} $\hat{m} = \hat{p}_{max} - \hat{p}_{(2)}$, where  $\hat{p}_{max}+ \hat{p}_{(2)}+\hat{p}_{min}=1$ and $\hat{p}_{(2)}$ is the second largest probability, or the \textit{entropy} of the predictions, a common measure in uncertainty quantification.

In practice, we propose the \BALToR{} (Bounded-Abstention Learning To Rank) procedure, which allows us to pass from a given ranker $f$ to an abstention mechanism over pairs (see \cref{alg:est0-1}).
We assume \textit{as given}: a ranker $f$, a calibration set $\mcD_{cal}$,  and a target coverage $c$. First, the algorithm estimates the conditional probabilities $\hat{p}_y\approx\pyxx$ for every $y\in\mathcal{Y}$ and stacks them in a vector $\hat{\mathbf{p}} = \{\hat{p}_{y,i}\}_{y\in\mathcal{Y}, i\in\mcD_{cal}}$ (line 1).
Then, the algorithm estimates the conditional risk by using one of the possible ways we mentioned above (i.e., the \textit{confidence}, the \textit{entropy} or the \textit{margin}) over each $i \in \mcD_{cal}$ (line 2). Next, it computes the $c-$th quantile $\hat{\beta}_c$ over the estimated conditional risks (line 3). The estimated quantile is used as a threshold in the returned selection function (line 4): if the estimated conditional risk of a pair of instances is below $\hat{\beta}_{c}$, the prediction is provided; otherwise, the selective ranker abstains.

We lastly need to explain how to actually estimate the conditional probabilities $\hat{p}_y\approx\pyxx$ starting from a ranker.
Indeed, given $s$, the learned ranking scores of each item, one can estimate $P(Y=1\mid \mbX=\mbx, \mbX'=\mbx')$, i.e., the probability that $\mbx$ is ranked higher than $\mbx'$, $P(Y=-1\mid \mbX=\mbx, \mbX'=\mbx')$, i.e., the probability that $\mbx$ is ranked lower than $\mbx'$, and $P(Y=0\mid \mbX=\mbx, \mbX'=\mbx')$, i.e., the two instances are tied, through parametric probability models~\citep{zhou2008learning}. The two most famous approaches are the \textit{Bradley-Terry model}~\citep{bradley1952rank} (BT) and the
\textit{Thurstone-Mosteller model }~\citep{thurstone1994law} (TM) (for tie-adjusted models, see, e.g., \cite{zhou2008learning}).
The former estimates probabilities as follows:
\begin{equation*}
    \hat{P}(Y=1\mid \mbX=\mbx, \mbX'=\mbx') = \frac{e^{s(\mbx)}}{e^{s(\mbx)}+\theta e^{s(\mbx')} }\; ,
\end{equation*}
\begin{equation*}
    \hat{P}(Y=-1\mid \mbX=\mbx, \mbX'=\mbx') = \frac{e^{s(\mbx')}}{e^{s(\mbx')}+\theta e^{s(\mbx)} }\; ,
\end{equation*}
\begin{multline*}
    \hat{P}(Y=0\mid \mbX=\mbx, \mbX'=\mbx') = \\\frac{(\theta-1)^2e^{s(\mbx)}e^{s(\mbx')}}{(\theta e^{s(\mbx)}+ e^{s(\mbx')})(e^{s(\mbx)}+\theta e^{s(\mbx')}) }\; ,
\end{multline*}
where $\theta= e^{\epsilon}$, with $\epsilon$ a parameter regulating the probabilities for ties.
The latter computes probabilities as follows:
\begin{equation*}
    \hat{P}(Y=1\mid \mbX=\mbx, \mbX'=\mbx')=\Phi\left(s(\mbx)-s(\mbx')-\epsilon\right)\; ,
\end{equation*}
\begin{equation*}
    \hat{P}(Y=-1\mid \mbX=\mbx, \mbX'=\mbx')=\Phi\left(s(\mbx')-s(\mbx)-\epsilon\right)\; ,
\end{equation*}
\[
\begin{split}
    \hat{P}(Y=0\mid \mbX=\mbx, \mbX'=\mbx')= \Phi\left(s(\mbx)-s(\mbx')+\epsilon\right)\\-\Phi\left(s(\mbx)-s(\mbx')-\epsilon\right)\; ,
\end{split}
\]
where $\Phi$ is the Gaussian cumulative distribution function.

Consequently, Bradley-Terry (BT) and Thurstone-Mosteller (TM) models allow us to estimate the desired conditional probabilities. Moreover, since (pairwise) learning to rank models often employ these frameworks to derive ranking scores $s$, it is natural to utilize these scores to recover the underlying conditional probabilities.

\IncMargin{1.2em}
\begin{algorithm2e}[!t]
\small 
    \caption{\BALToR{}}
    \label{alg:est0-1}
	\SetKwInOut{Input}{Input}
	\SetKwInOut{Output}{Output}
	\Input{$\mathcal{D}_{cal}, f,c$ -- held-out (un)labeled dataset $\mathcal{D}_{cal}$, trained ranker $f$, target coverage $c$
    }
	\Output{$g$ - selection function}
	\BlankLine
        $\hat{\mathbf{p}}  \leftarrow \{f.rank\_proba((\mbx,\mbx')_i)\}_{i\in\mathcal{D}_{cal}}$
        \tcp*[f]{conditional probabilities on $\mathcal{D}_{cal}$}\\
        $\hat{\mathbf{r}}_{cal} \leftarrow \{estimated\_risk(\mbx,\mbx')_i \}_{i\in\mathcal{D}_{cal}}$\hspace{10ex}\tcp*[f]{estimate conditional risk on $\mathcal{D}_{cal}$}\\
        $\hat{\beta}_c \leftarrow quantile(\hat{\mathbf{r}}_{cal}, c)$\hspace{25ex}\tcp*[f]{$c$-th quantile estimate}\\ 
        $g\leftarrow \text{\textbf{lambda}}(\mbx,\mbx'):\mathbb{I}_{(1-\max{f.rank\_proba\left((\mbx,\mbx')\right)} < \hat{\beta}_c)}$\tcp*[f]{selection function}\\
    \Return{$g$}
\end{algorithm2e}

\section{Experiments}
\label{sec:experiments}

\begin{figure*}[t]
\centering
\begin{subfigure}[t]{0.24\textwidth}
\centering
    \includegraphics[scale=.145]{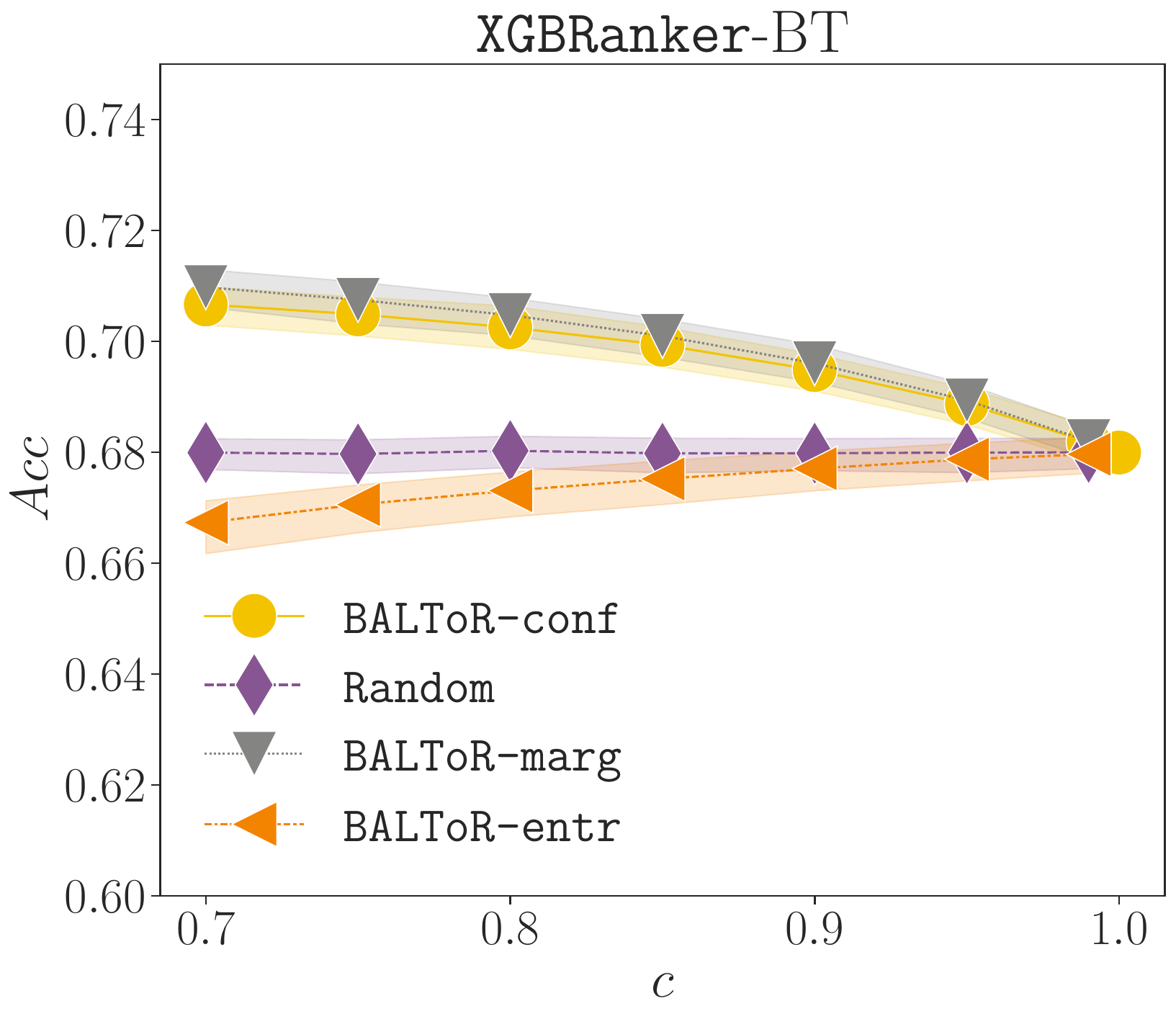}
    \caption{\texttt{MQ2007}}
    \label{fig:BTMQ2007}
\end{subfigure}
\hfill
\begin{subfigure}[t]{0.24\textwidth}
        \centering
    \includegraphics[scale=.145]{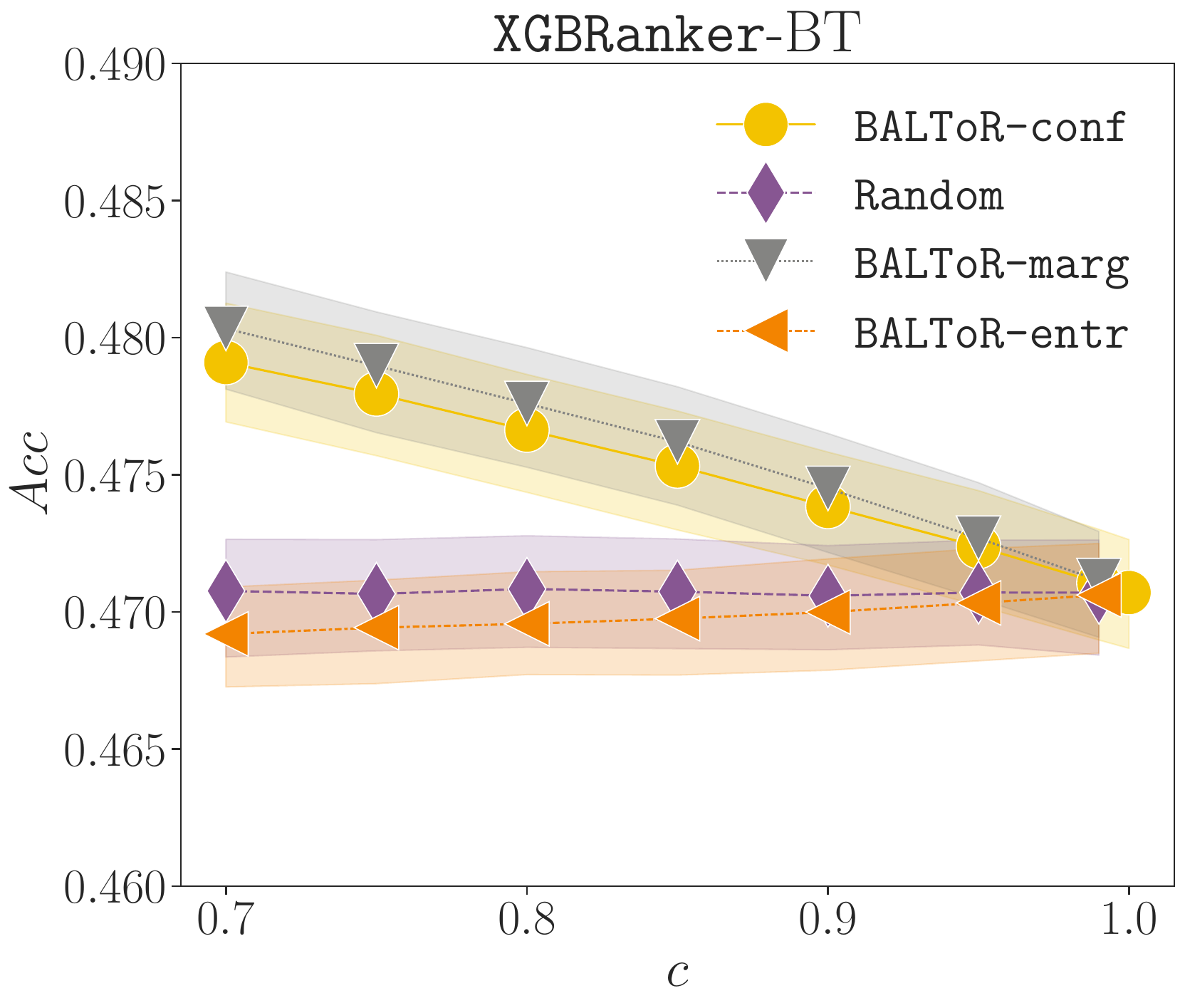}
    \caption{\texttt{Web-30k}}
    \label{fig:BTMSLR}
\end{subfigure}
\hfill
\begin{subfigure}[t]{0.24\textwidth}
        \centering
    \includegraphics[scale=.145]{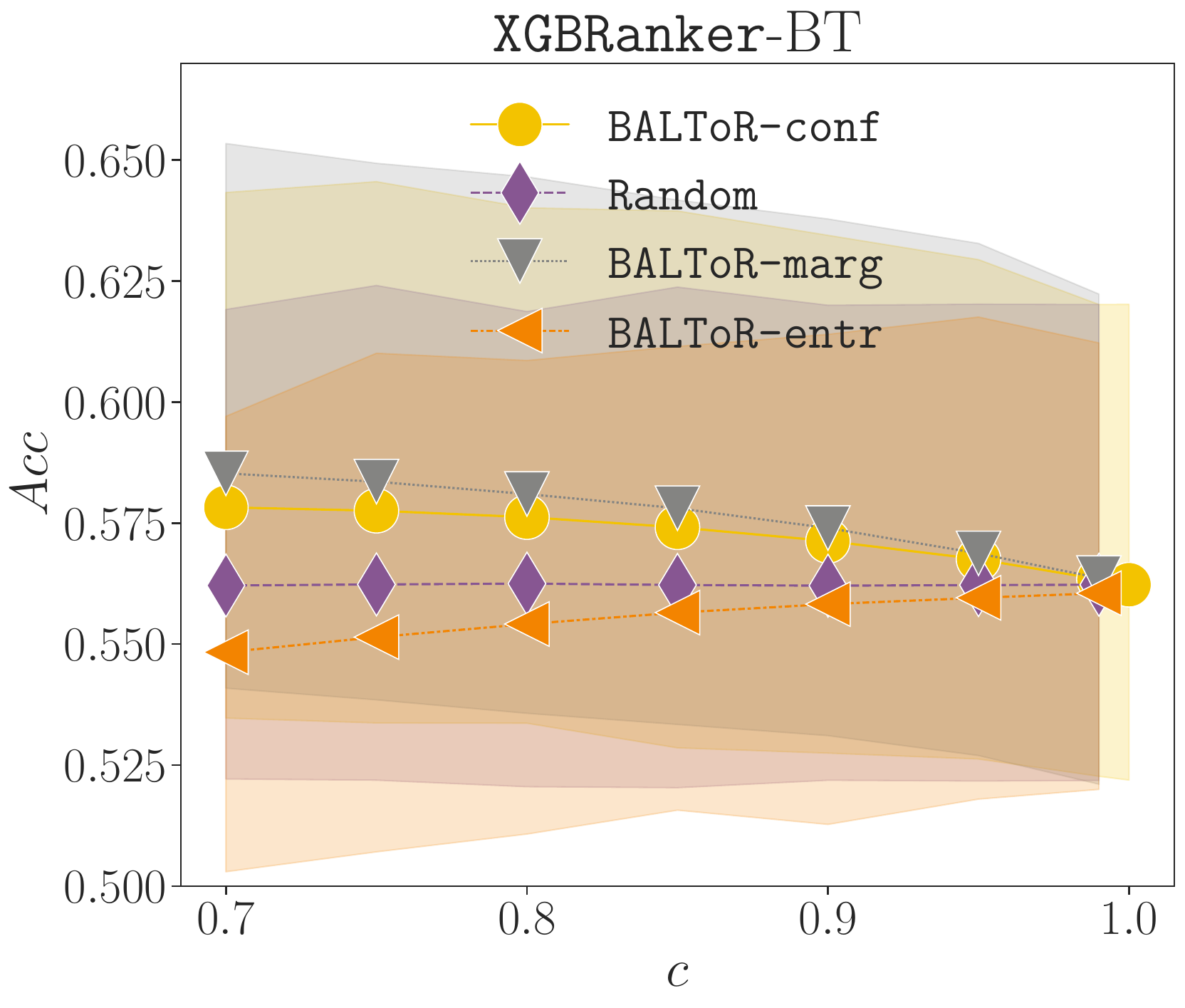}
    \caption{\texttt{OHSUMED}}
    \label{fig:BTOHS}
\end{subfigure}
\hfill
\begin{subfigure}[t]{0.24\textwidth}
        \centering
    \includegraphics[scale=.145]{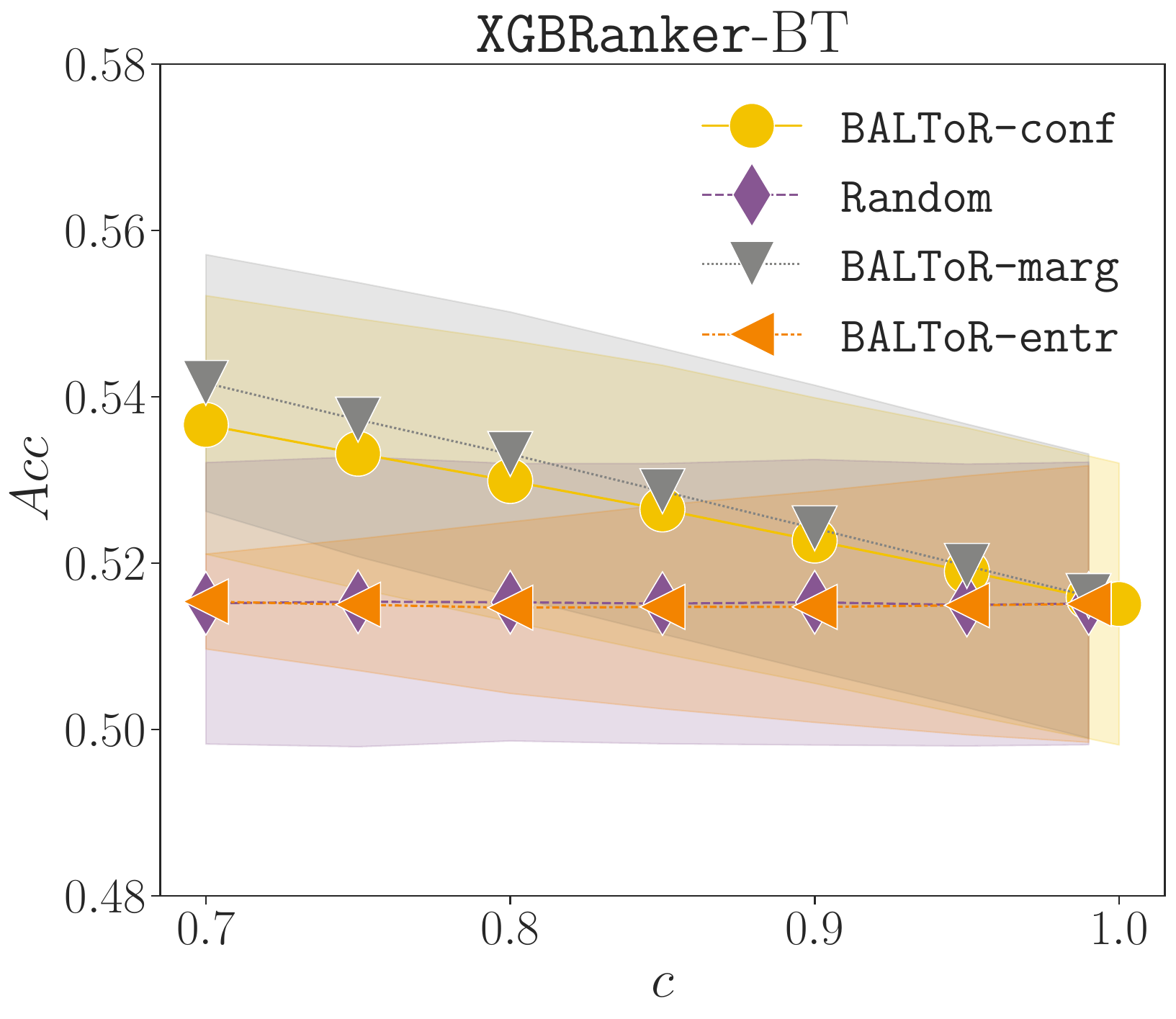}
    \caption{\texttt{Yahoo}}
    \label{fig:BTYAH}
\end{subfigure}

\caption{Accuracy $Acc$ on the selected pairs (mean and $95\%$ confidence intervals) over default folds for the BT model. For smaller target coverages $c$ (i.e., more abstention), the accuracy increases for \BALToRc{} and \BALToRm{}, remains stable for the random abstainer, and has erratic performance for \BALToRe{}.}
\label{fig:AccResBT}
\end{figure*}

\begin{figure*}[t]
\centering
\begin{subfigure}[t]{0.24\textwidth}
\centering
    \includegraphics[scale=0.145]{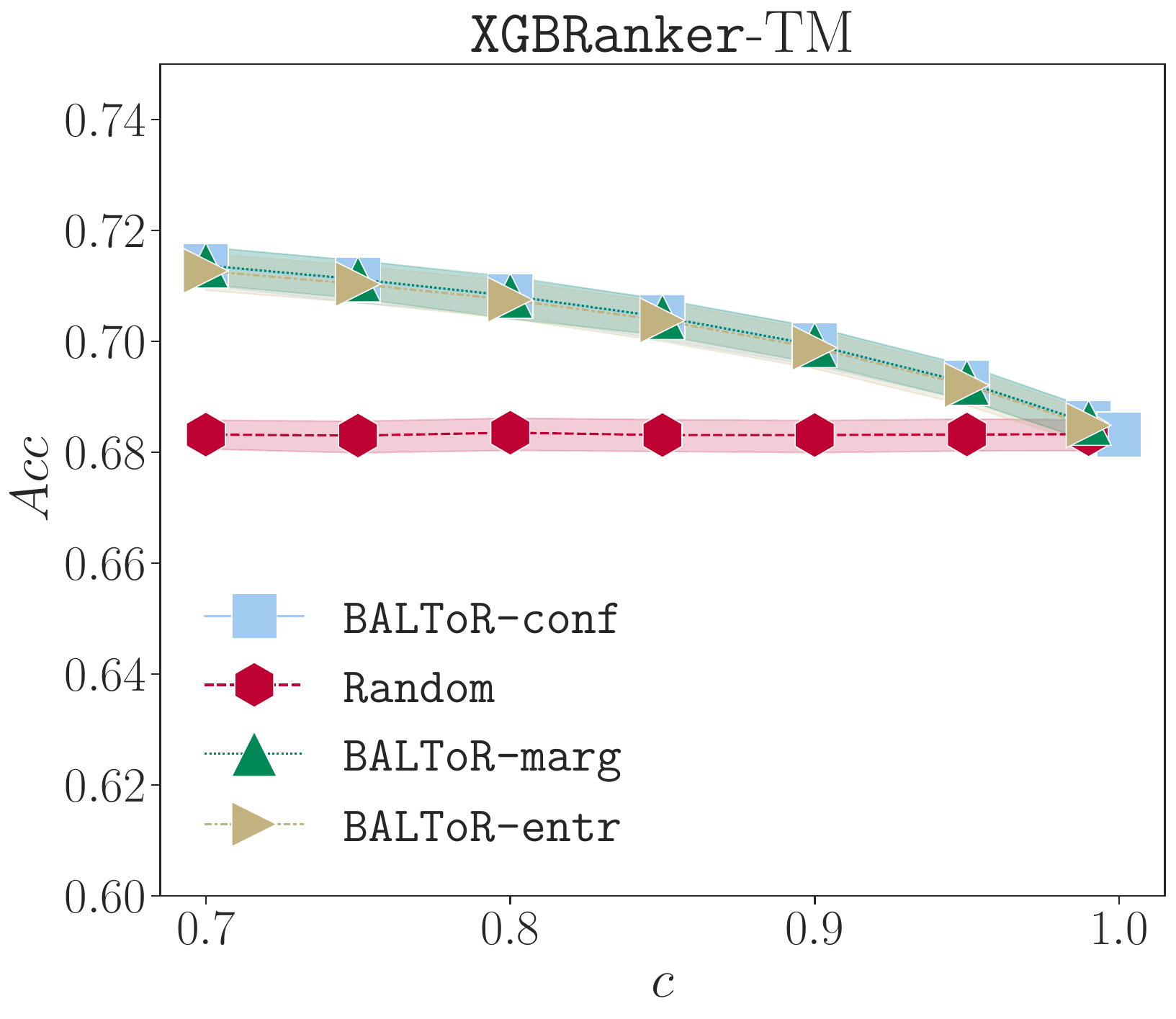}
    \caption{\texttt{MQ2007}}
    \label{fig:TMMQ2007}
\end{subfigure}
\hfill
\begin{subfigure}[t]{0.24\textwidth}
        \centering
    \includegraphics[scale=0.145]{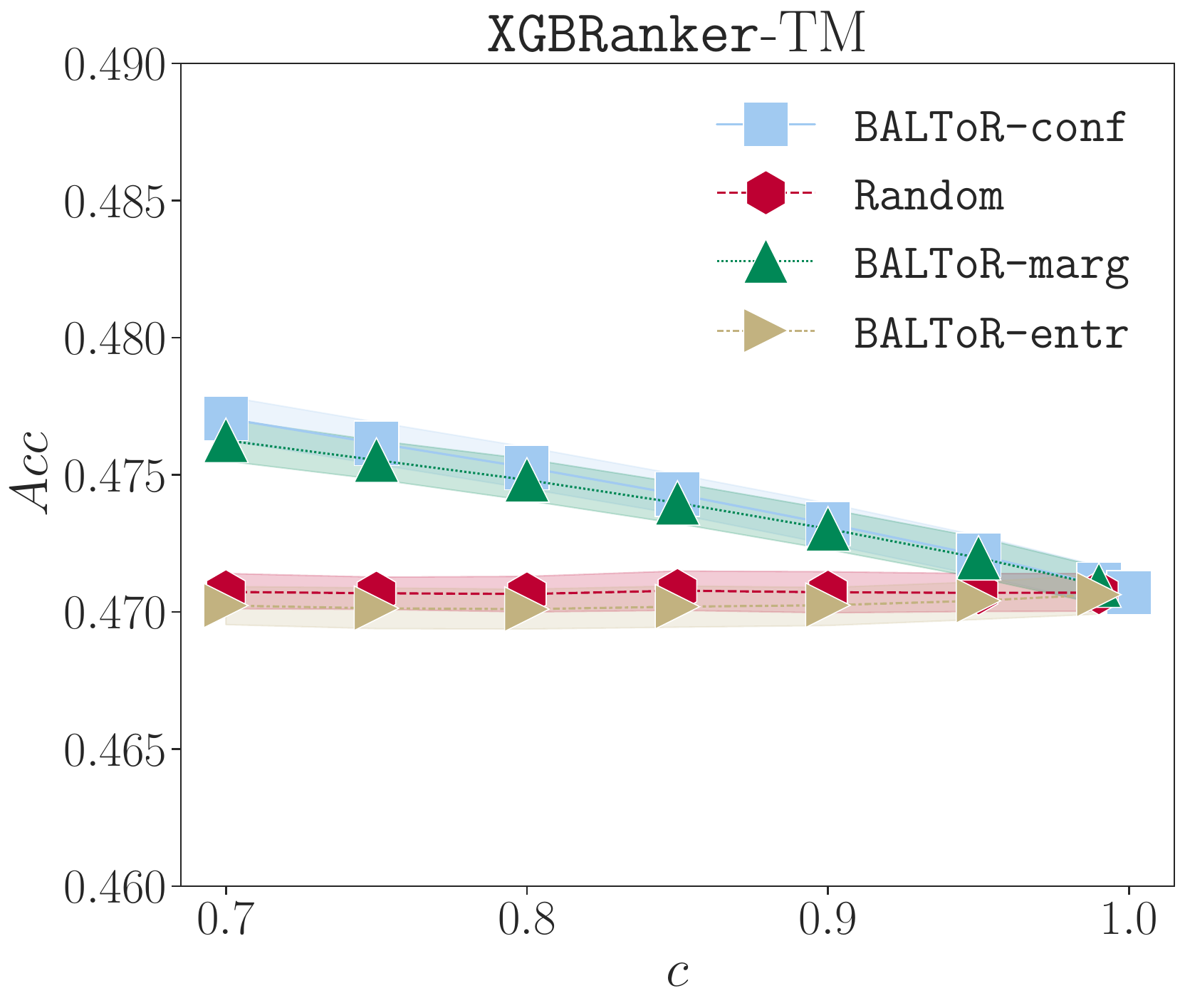}
    \caption{\texttt{Web-30k}}
    \label{fig:TMMSLR}
\end{subfigure}
\hfill
\begin{subfigure}[t]{0.24\textwidth}
        \centering
    \includegraphics[scale=0.145]{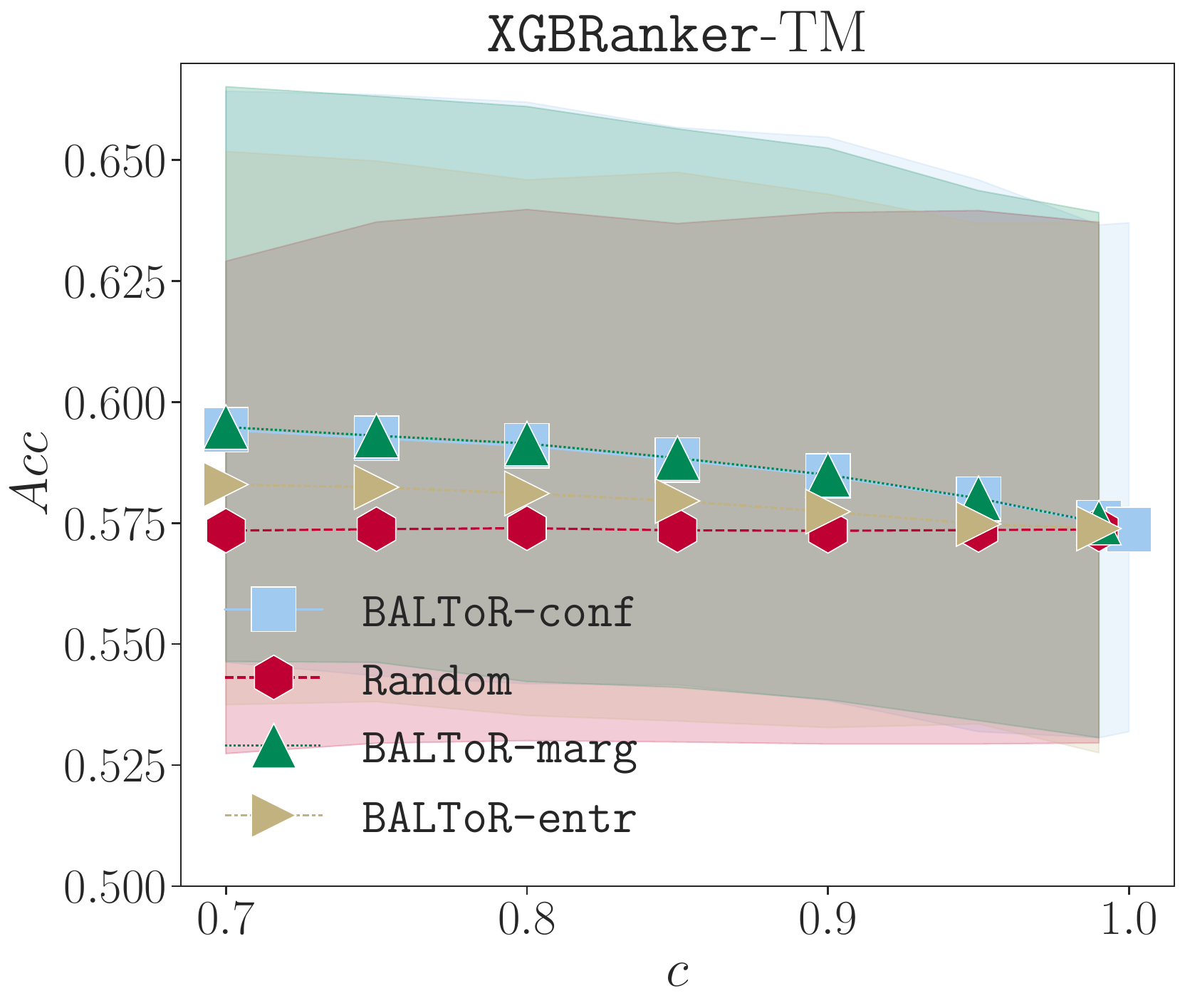}
    \caption{\texttt{OHSUMED}}
    \label{fig:TMOHS}
\end{subfigure}
\hfill
\begin{subfigure}[t]{0.24\textwidth}
        \centering
    \includegraphics[scale=0.145]{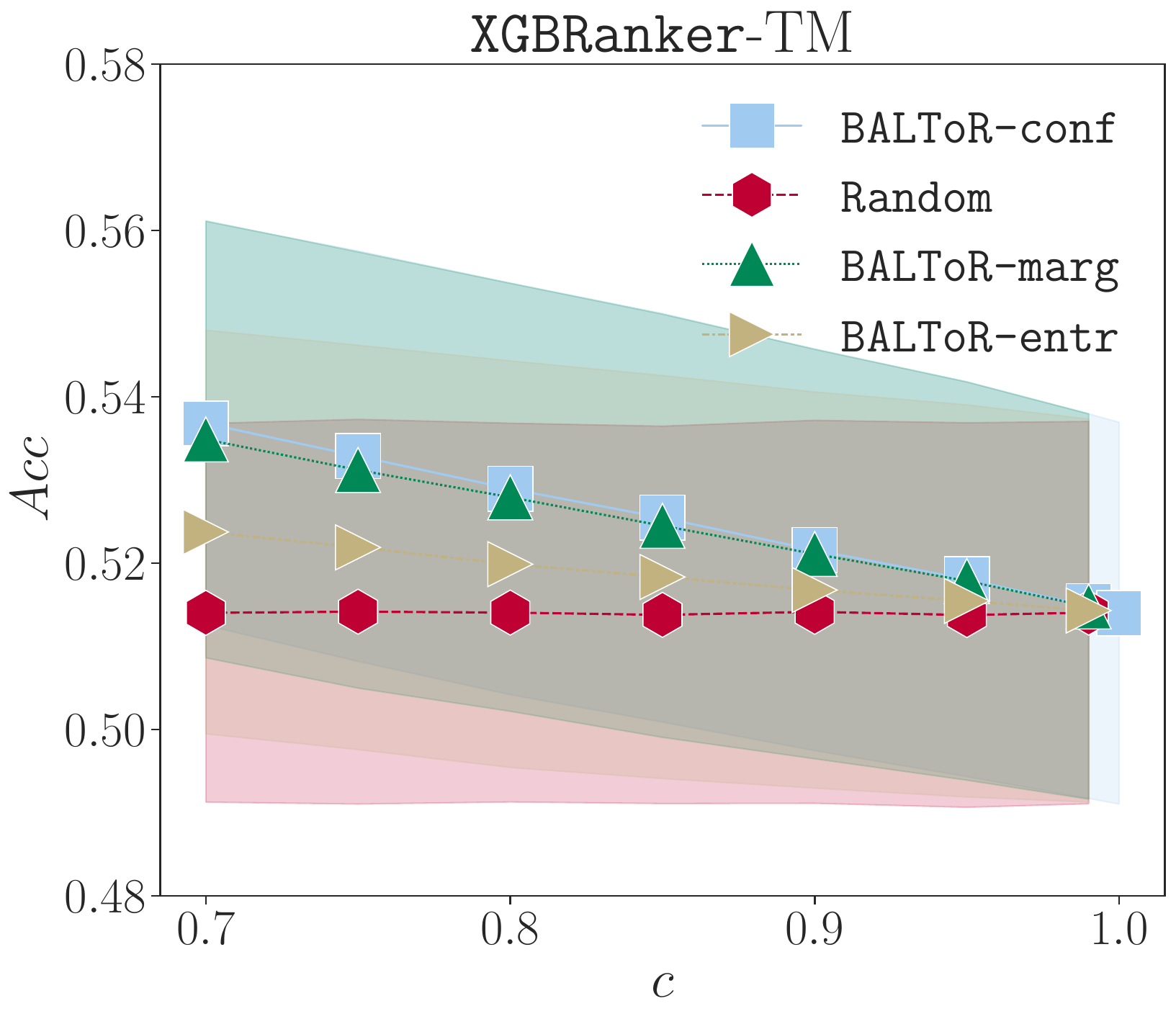}
    \caption{\texttt{Yahoo}}
    \label{fig:TMYAH}
\end{subfigure}

\caption{Accuracy $Acc$ on the selected pairs (mean and $95\%$ confidence intervals) over default folds for the TM model. For smaller target coverages $c$ (i.e., more abstention), the accuracy increases for \BALToRc{} and \BALToRm{}, remains stable for the random abstainer, and has erratic performance for \BALToRe{}.}
\label{fig:AccResTM}
\end{figure*}

\begin{figure*}[t]
    \centering
\begin{subfigure}[t]{00.48\textwidth}
    \includegraphics[scale=.30]{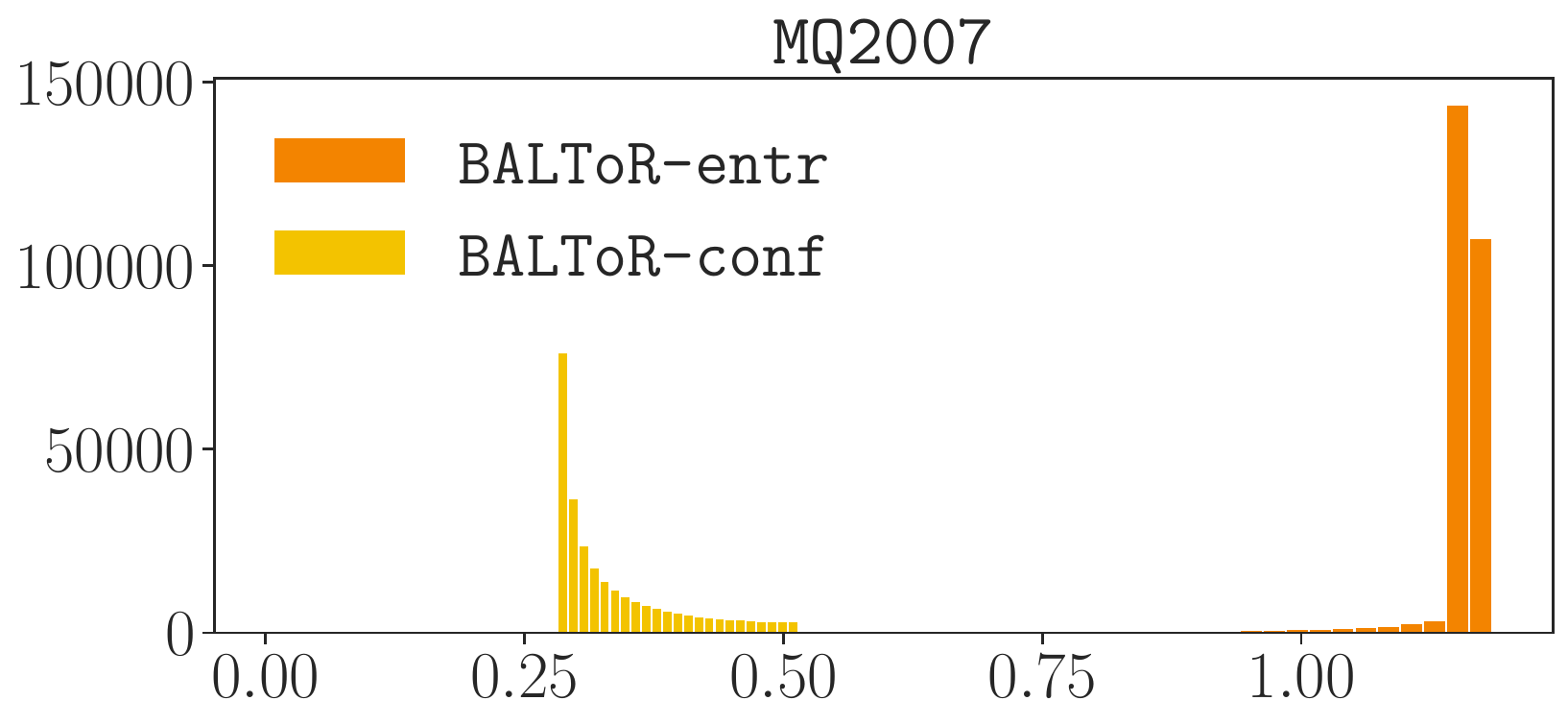}
    \caption{BT model}
    \label{fig:BTMQ2007_xgb_entropy}
\end{subfigure}
\hfill
\begin{subfigure}[t]{00.48\textwidth}
    \includegraphics[scale=.30]{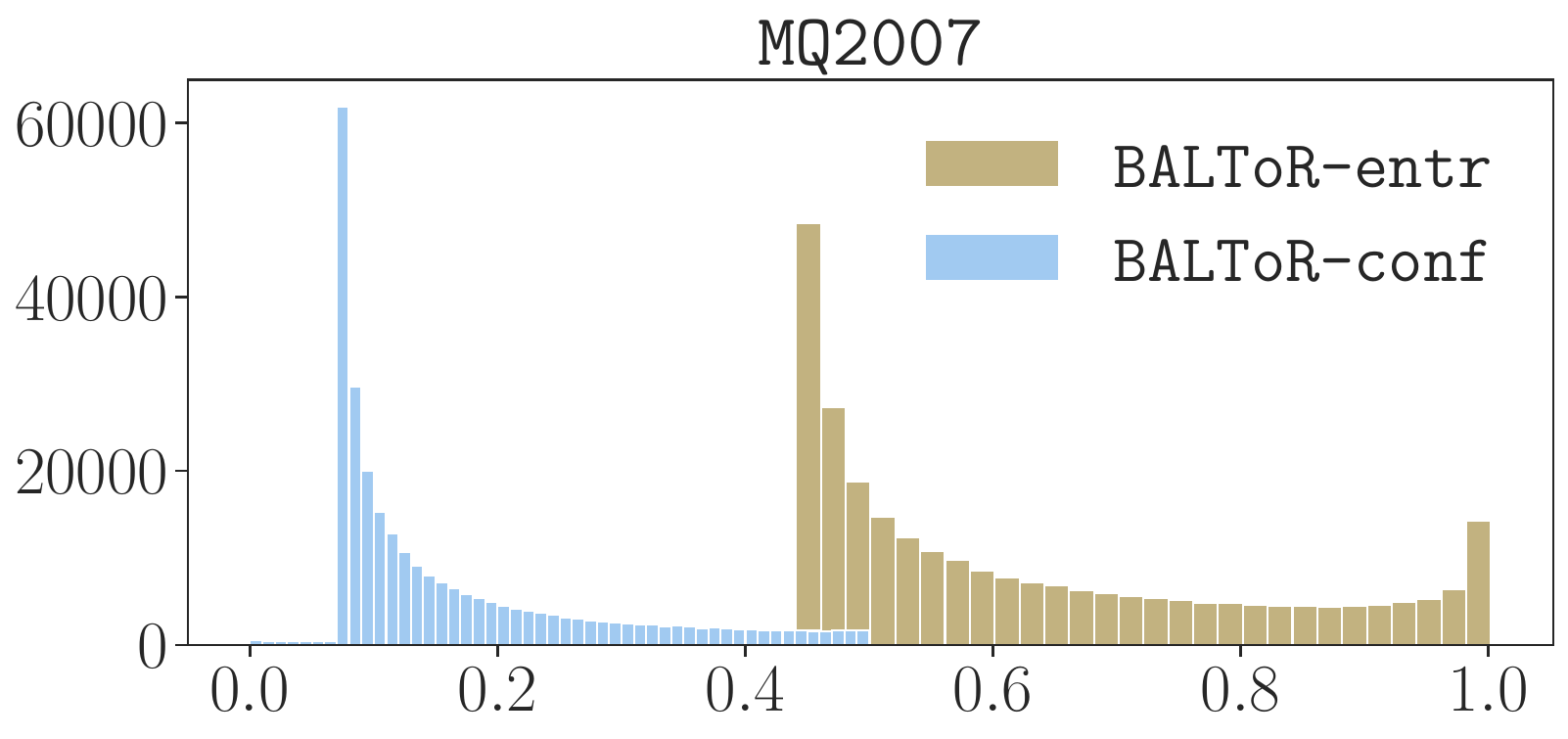}
    \caption{TM model}
    \label{fig:TMMQ2007_xgb_entropy}
\end{subfigure}
\caption{Estimated density functions over \texttt{MQ2007} Fold 1 calibration set for BT (\cref{fig:BTMQ2007_xgb_entropy}) and TM (\cref{fig:TMMQ2007_xgb_entropy}) when using an XGBRanker. The shapes of the density functions differ (are similar) when using BT (TM) model.}
\label{fig:example_entropy}
\end{figure*}

\begin{figure*}[t]
\centering
\begin{subfigure}[t]{0.24\textwidth}
    \includegraphics[scale=0.14]{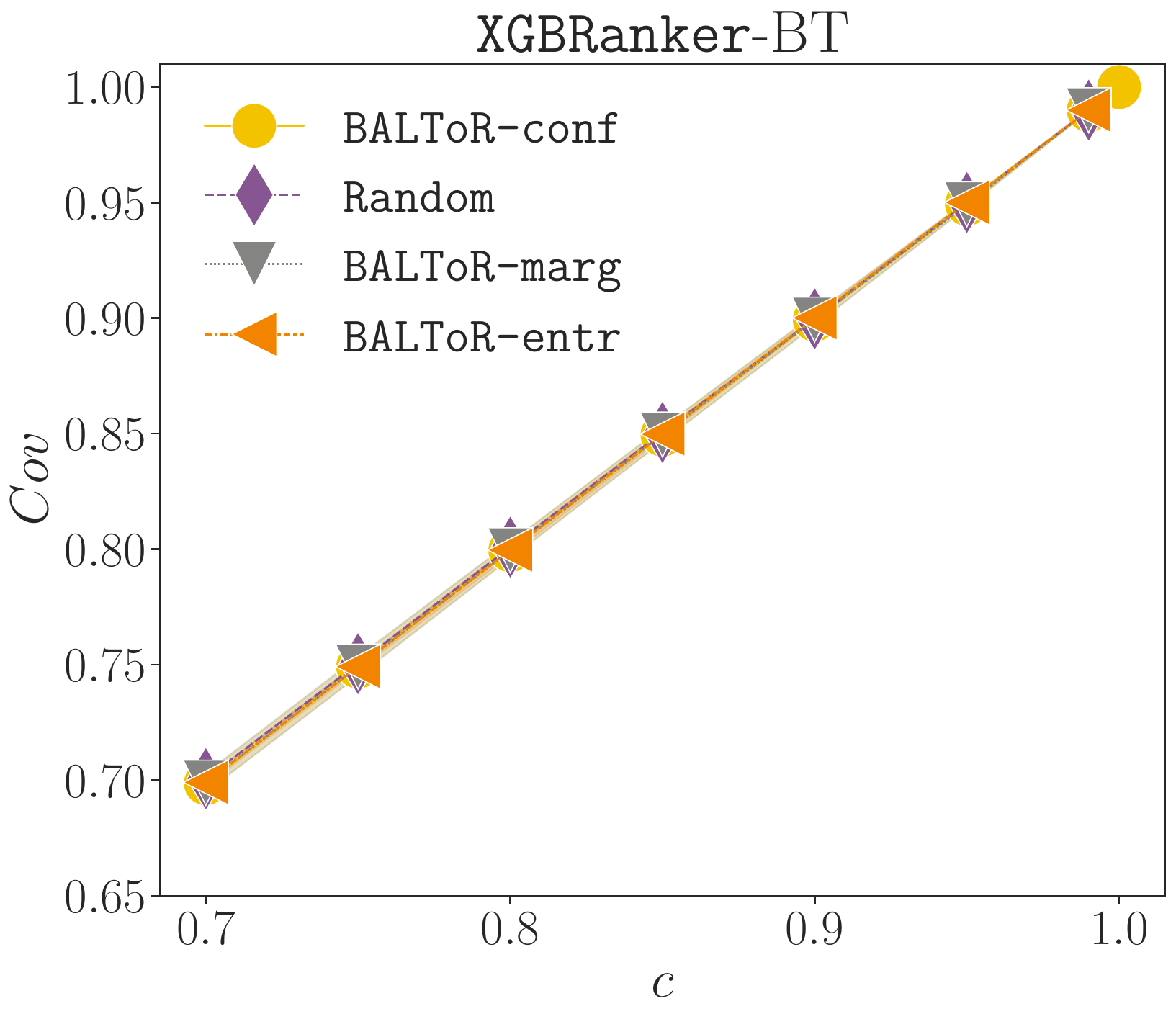}
    \caption{\texttt{MQ2007}}
    \label{fig:BTMQ_cov}
\end{subfigure}
\hfill
\begin{subfigure}[t]{0.24\textwidth}
    \includegraphics[scale=0.14]{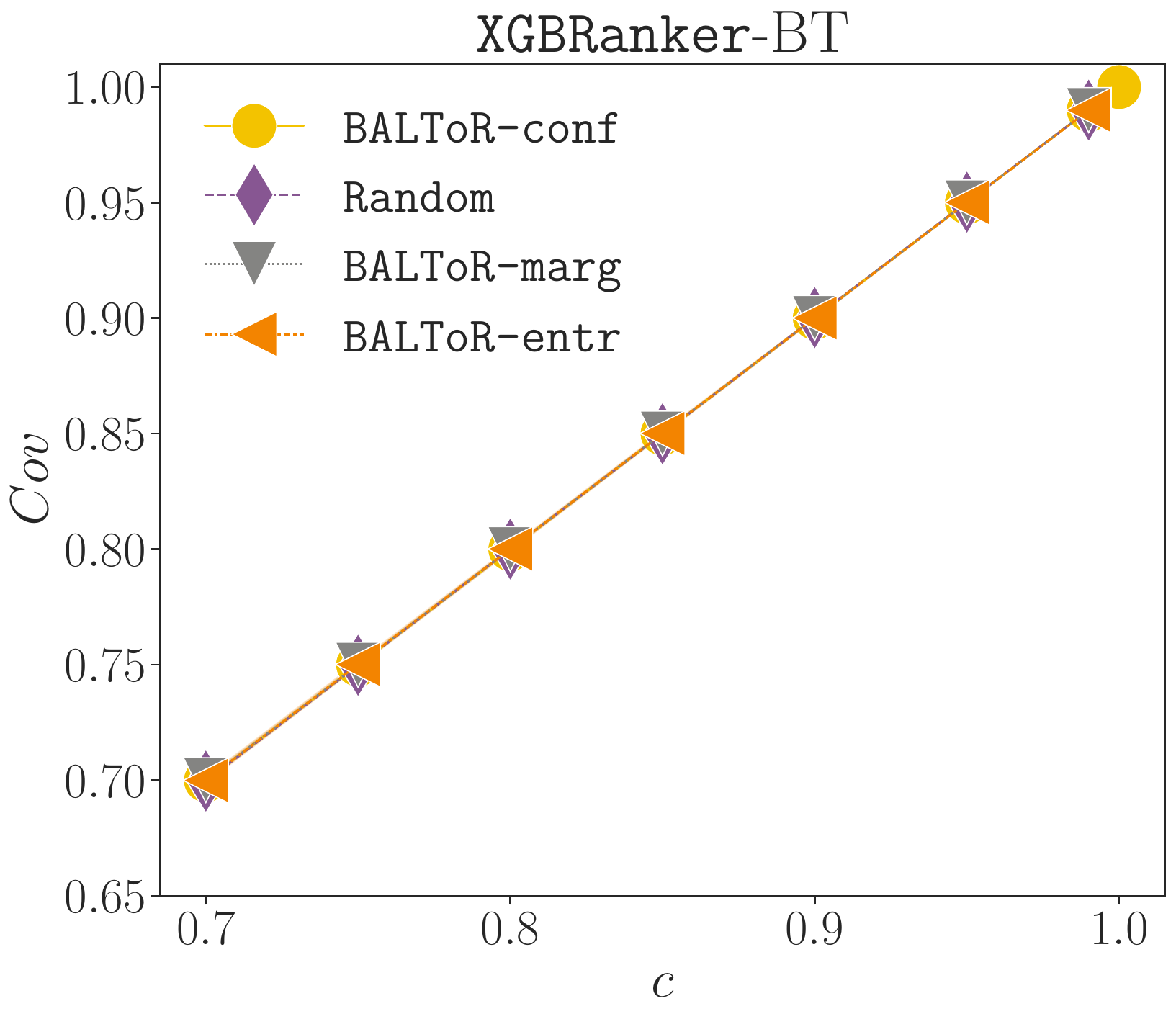}
    \caption{\texttt{Web-30k}}
    \label{fig:BTMSL_cov}
\end{subfigure}
\hfill
\begin{subfigure}[t]{0.24\textwidth}
    \includegraphics[scale=0.14]{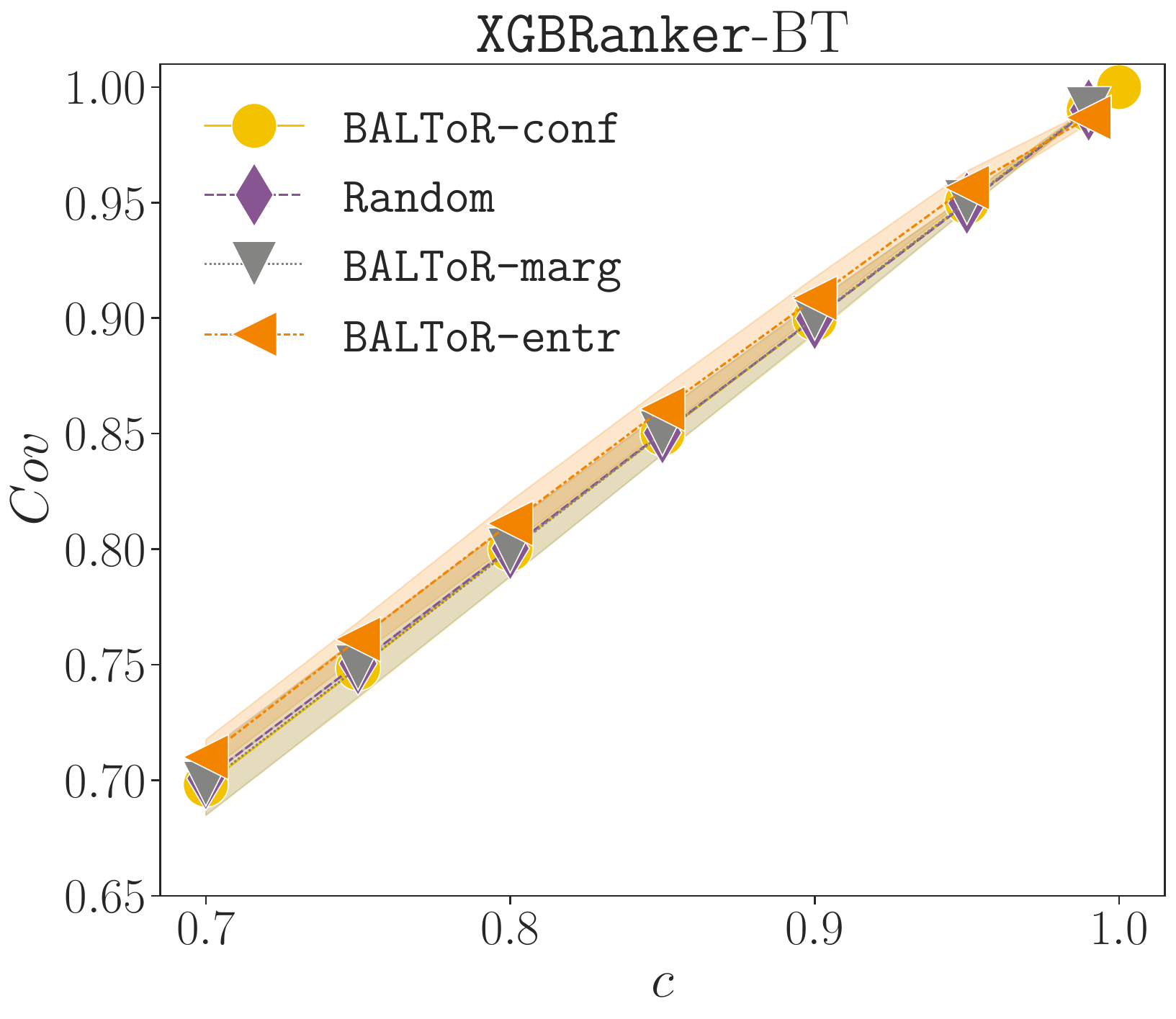}
    \caption{\texttt{OHSUMED}}
    \label{fig:BTOHS_cov}
\end{subfigure}
\hfill
\begin{subfigure}[t]{0.24\textwidth}
    \includegraphics[scale=0.14]{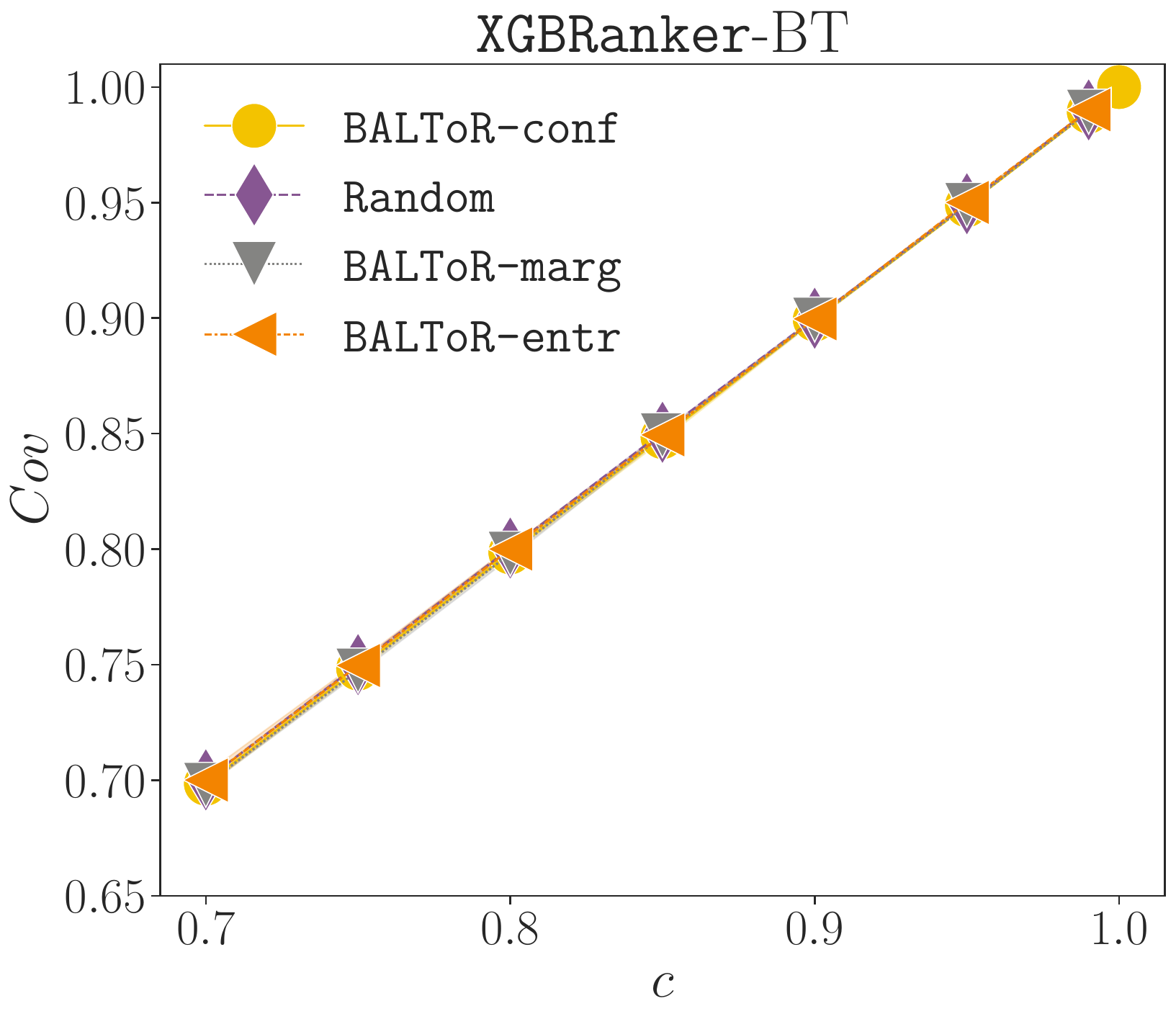}
    \caption{\texttt{Yahoo}}
    \label{fig:BTYAH_cov}
\end{subfigure}
\caption{Empirical coverage $Cov$ on the test set (mean and 95\% confidence intervals) for the BT model
. 
}
\label{fig:COVres}
\end{figure*}

\begin{figure*}[t]
\centering
\begin{subfigure}[t]{0.24\textwidth}
    \includegraphics[scale=0.14]{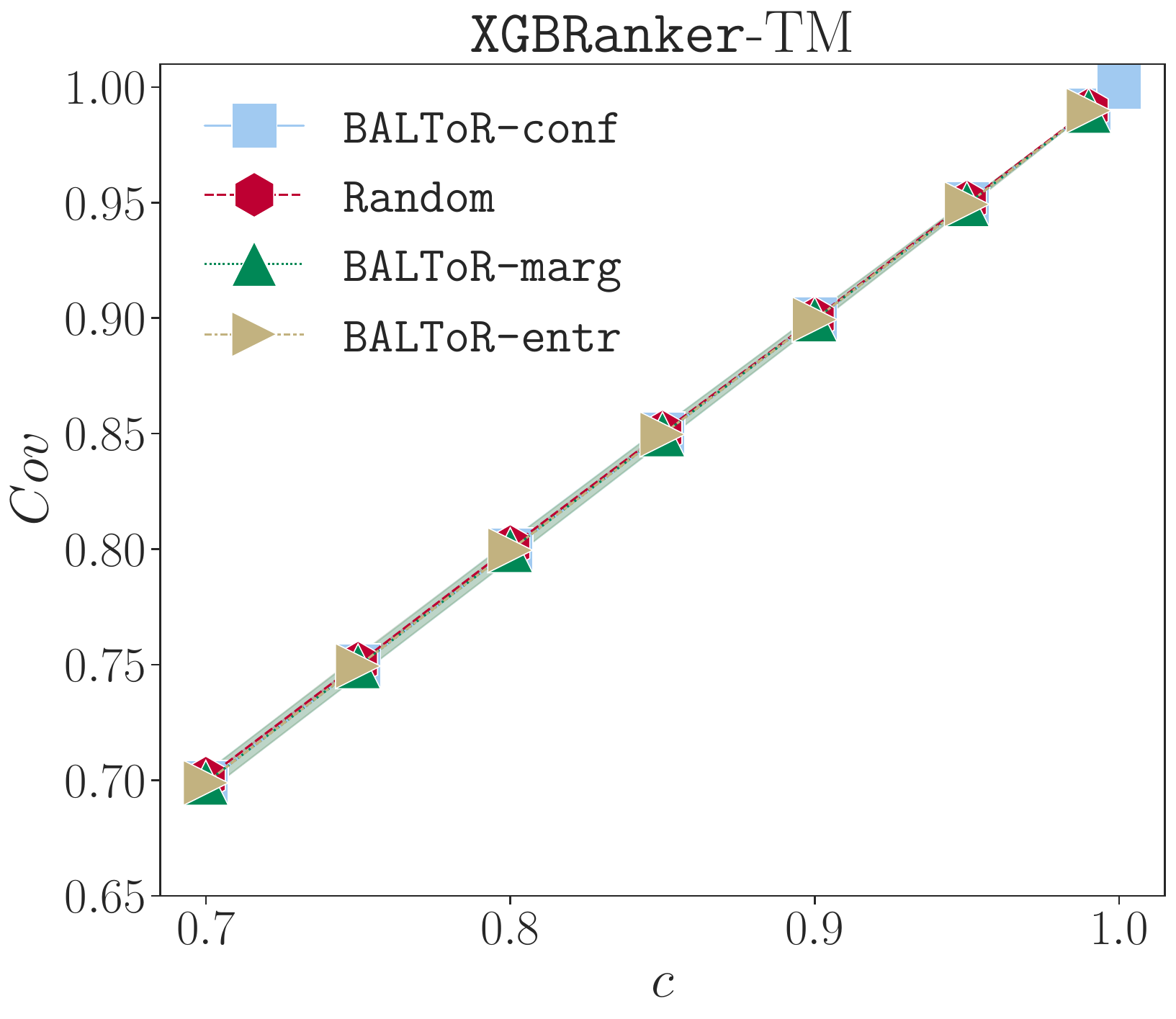}
    \caption{\texttt{MQ2007}}
    \label{fig:TMMQ_cov}
\end{subfigure}
\hfill
\begin{subfigure}[t]{0.24\textwidth}
    \includegraphics[scale=0.14]{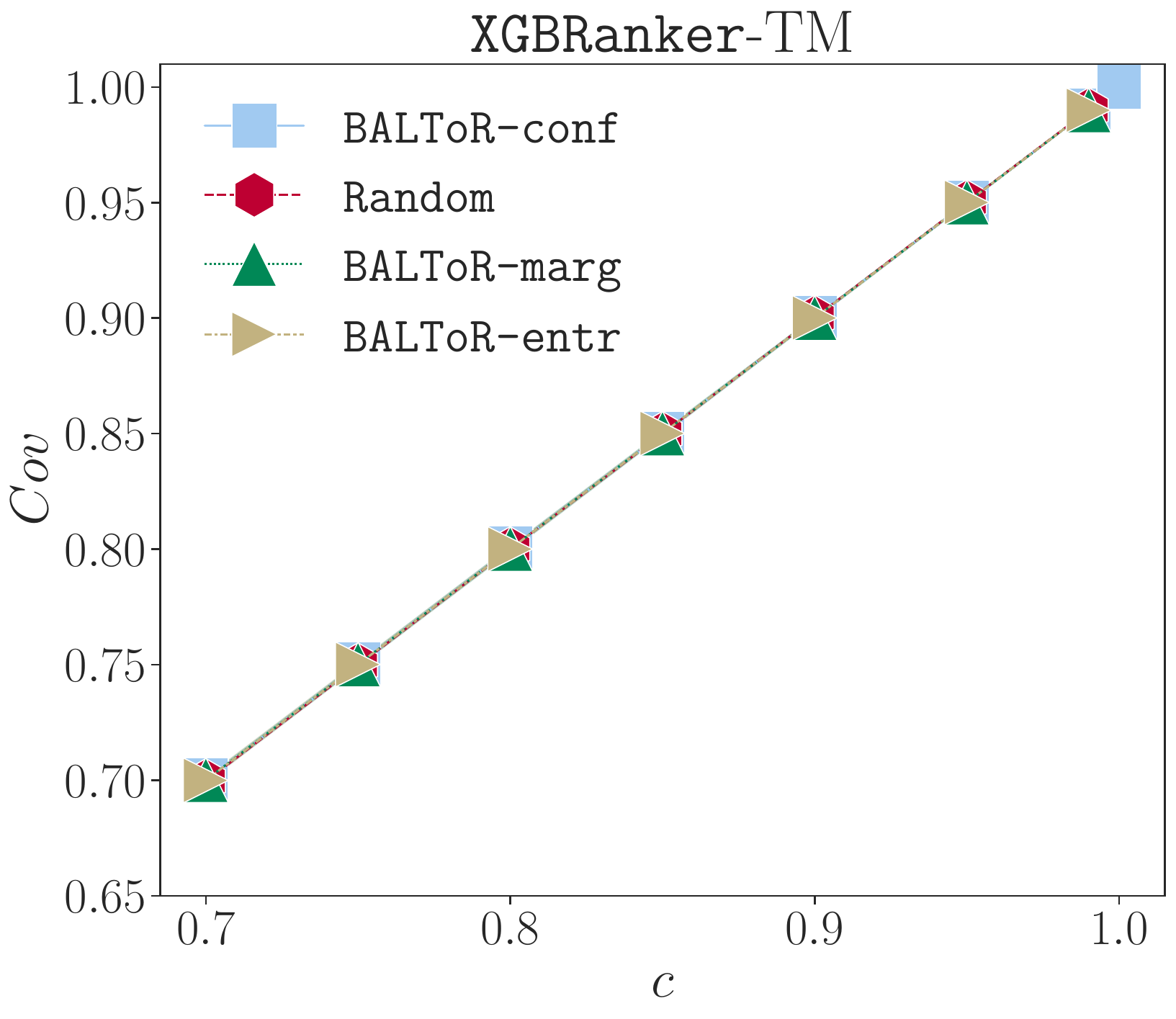}
    \caption{\texttt{Web-30k}}
    \label{fig:TMMSL_cov}
\end{subfigure}
\hfill
\begin{subfigure}[t]{0.24\textwidth}
    \includegraphics[scale=0.14]{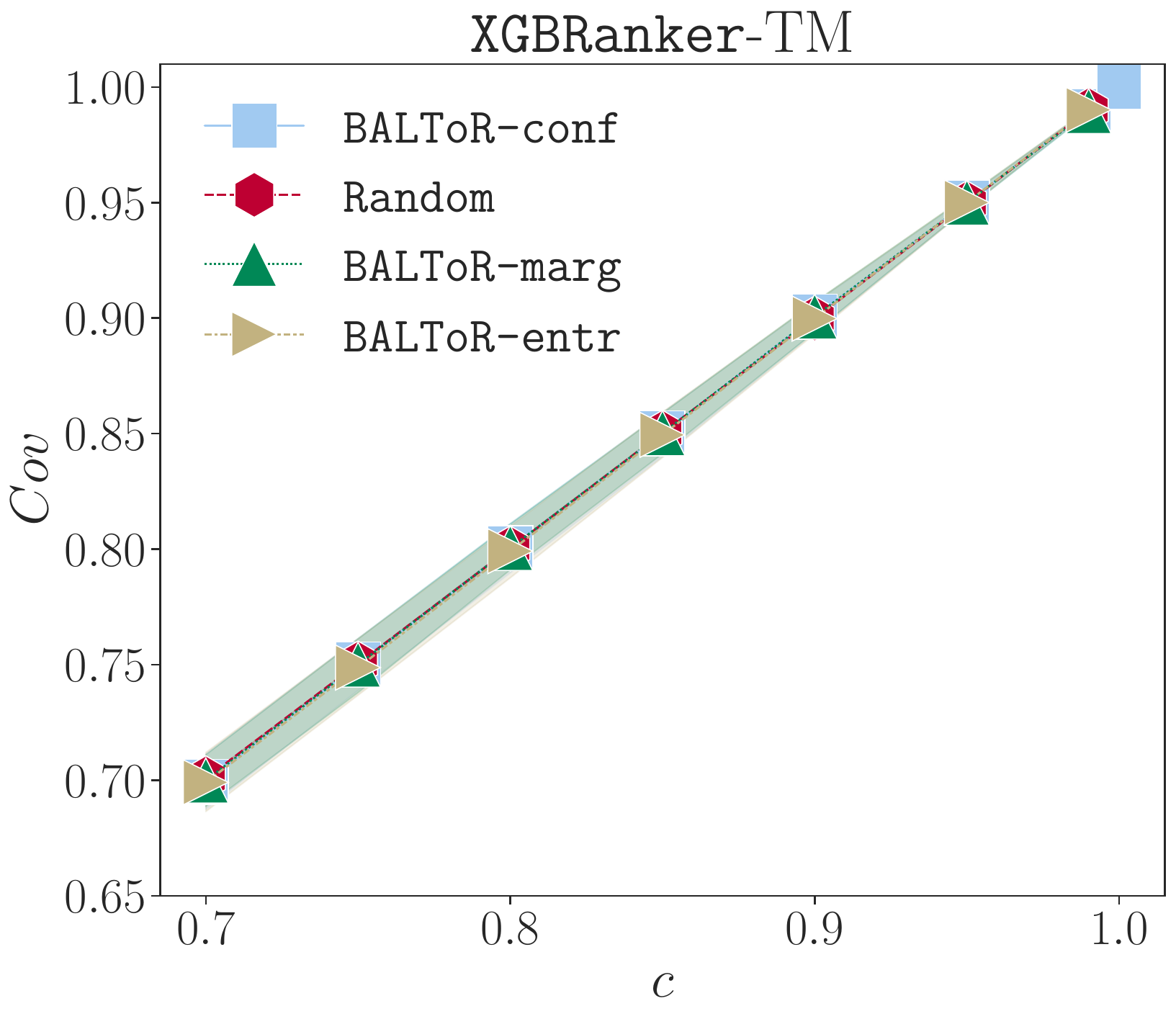}
    \caption{\texttt{OHSUMED}}
    \label{fig:TMOHS_cov}
\end{subfigure}
\hfill
\begin{subfigure}[t]{0.24\textwidth}
    \includegraphics[scale=0.14]{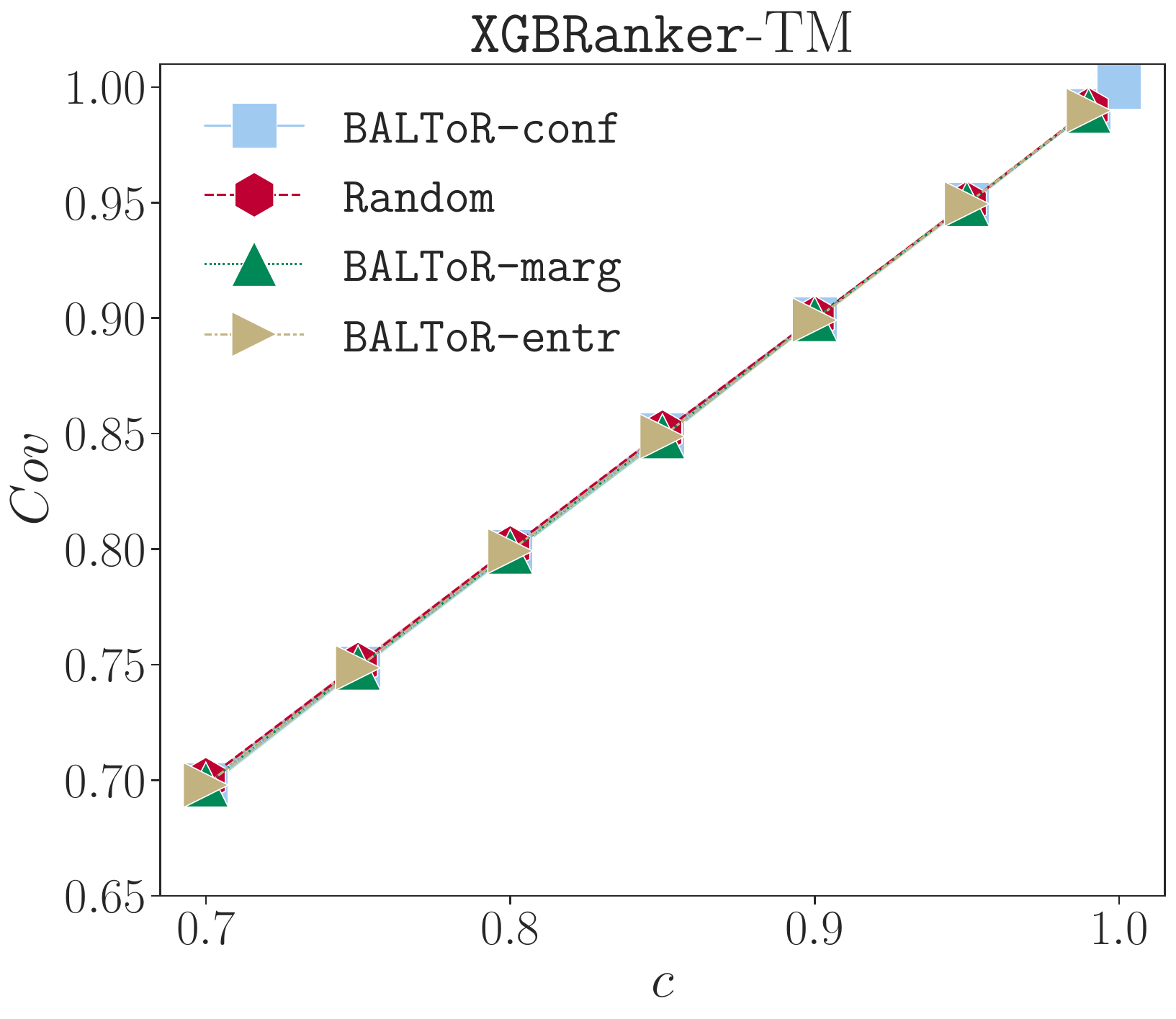}
    \caption{\texttt{Yahoo}}
    \label{fig:TMYAH_TMcov}
\end{subfigure}
\caption{Empirical coverage $Cov$ on the test set (mean and 95\% confidence intervals) for the TM model. 
}
\label{fig:COVresTM}
\end{figure*}

\begin{figure*}[t]
\centering
\begin{subfigure}[t]{0.24\textwidth}
    \includegraphics[scale=0.14]{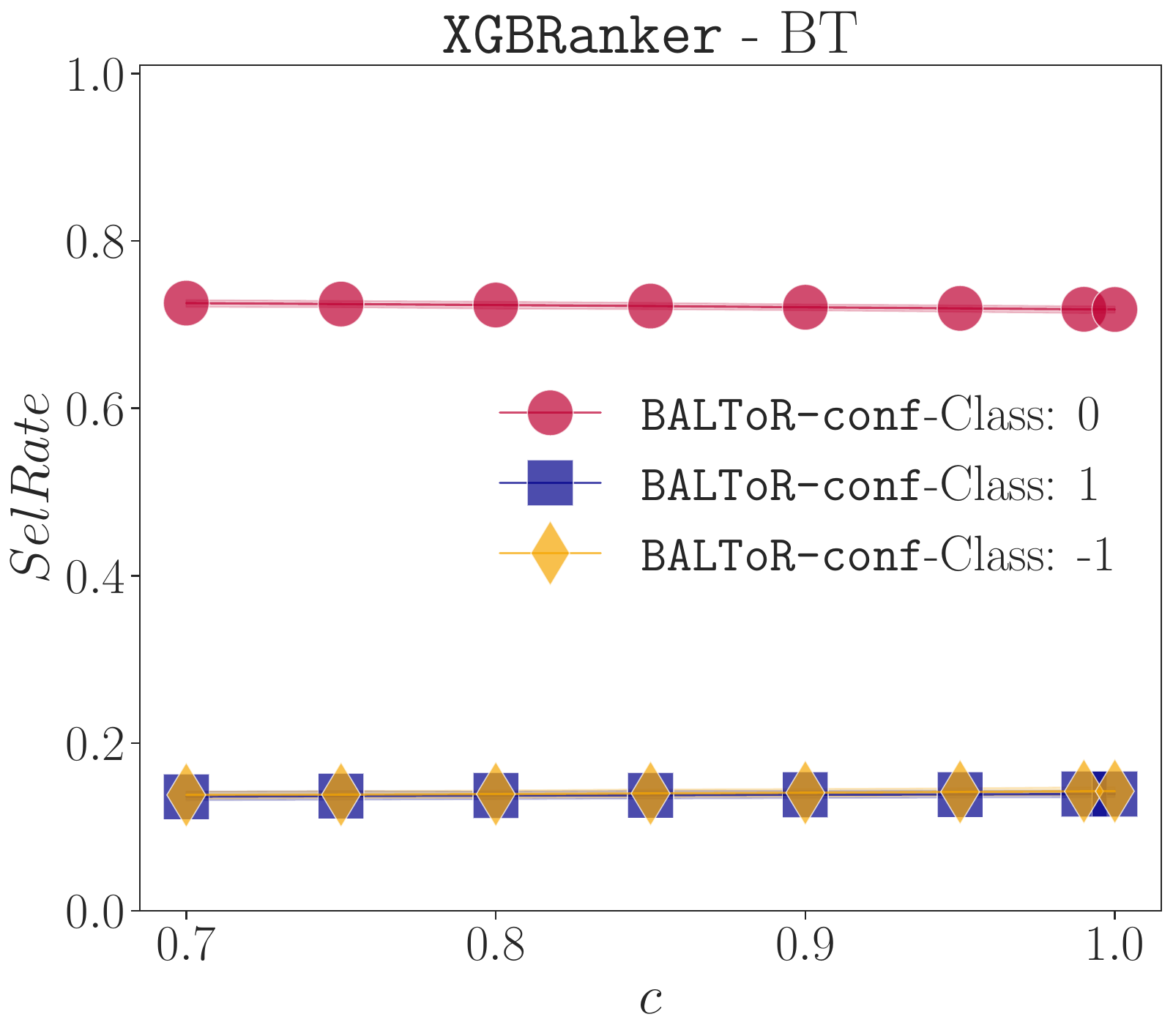}
    \caption{\texttt{MQ2007}}
    \label{fig:S1res_mq2007_BT}
\end{subfigure}
\hfill
\begin{subfigure}[t]{0.24\textwidth}
        \centering
    \includegraphics[scale=0.14]{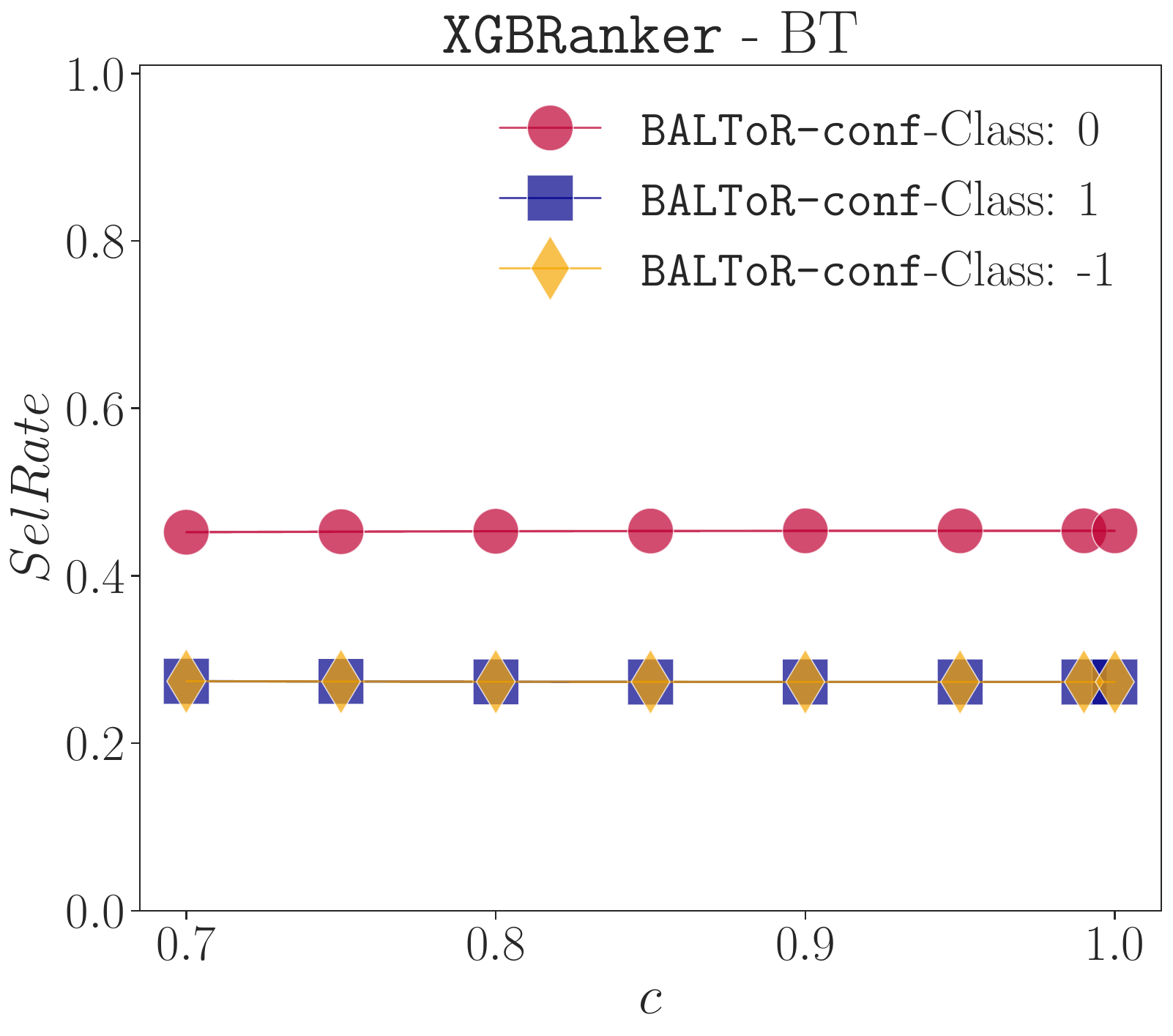}
    \caption{\texttt{Web-30k}}
    \label{fig:S1res_mslr_BT}
\end{subfigure}
\hfill
\begin{subfigure}[t]{0.24\textwidth}
        \centering
    \includegraphics[scale=0.14]{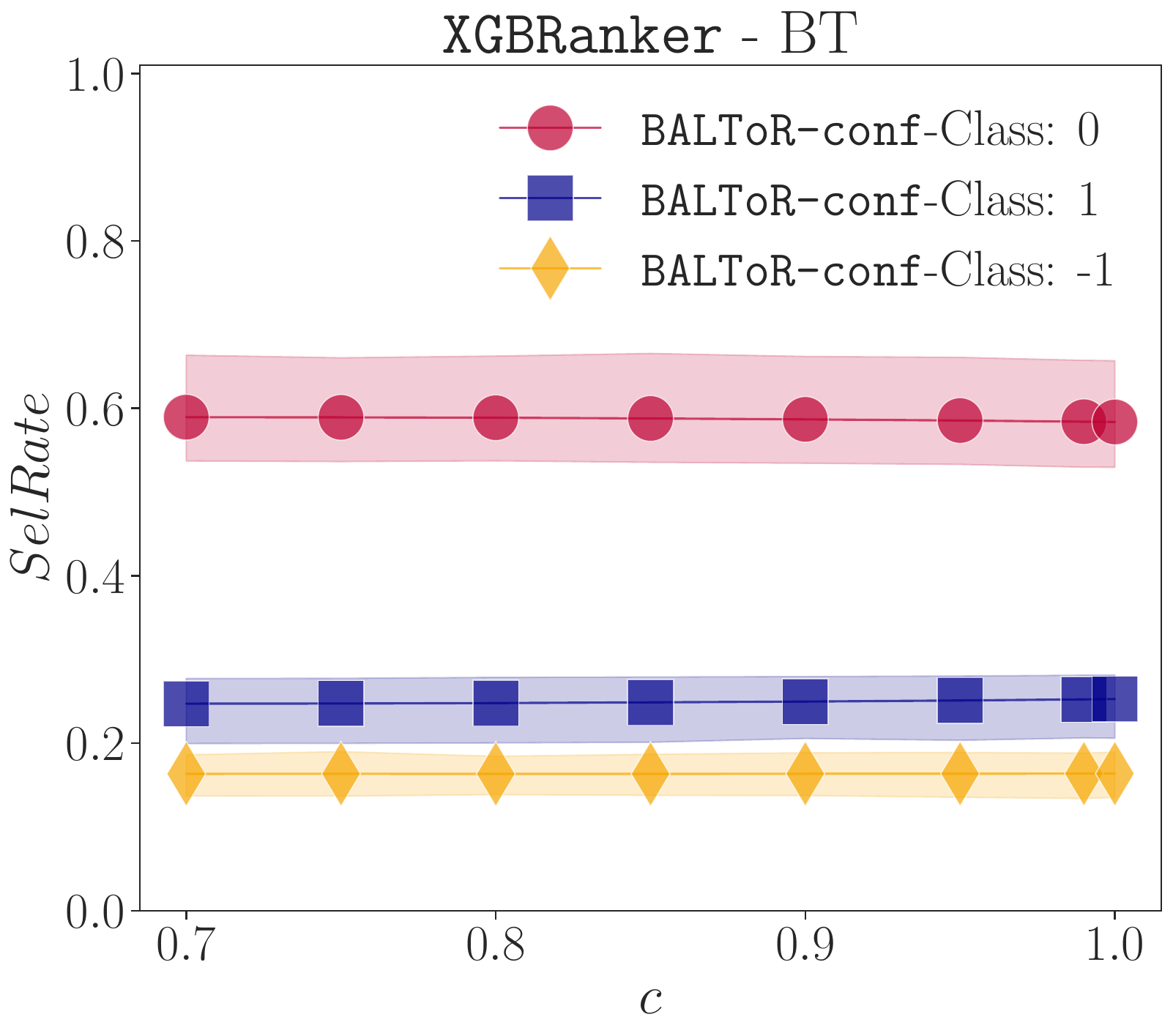}
    \caption{\texttt{OHSUMED}}
    \label{fig:S1res_ohsumed_BT}
\end{subfigure}
\hfill
\begin{subfigure}[t]{0.24\textwidth}
        \centering
    \includegraphics[scale=0.14]{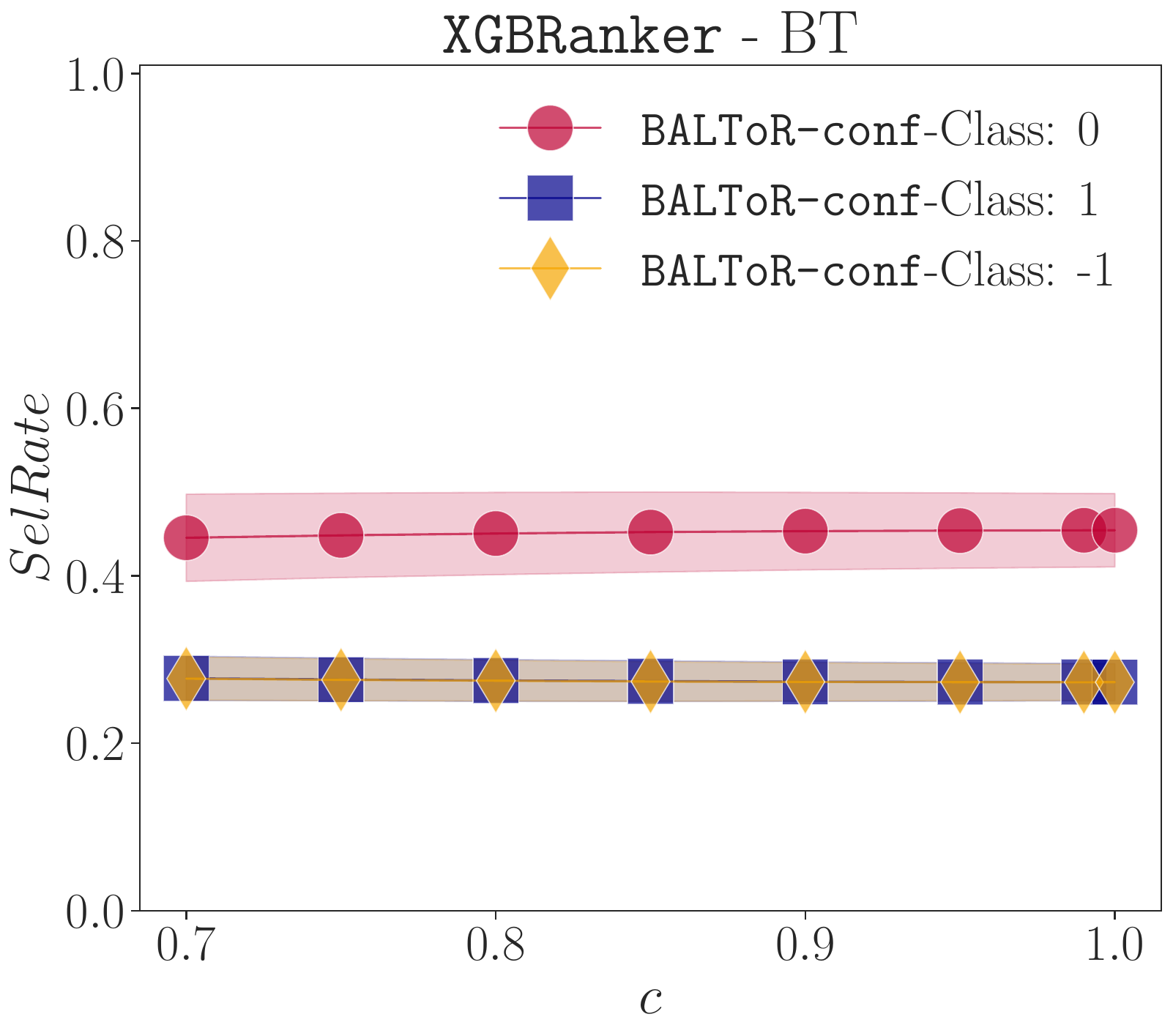}
    \caption{\texttt{Yahoo}}
    \label{fig:S1res_yahoo_BT}
\end{subfigure}
\caption{Class distribution $SelRate$ on the selected pairs (mean and 95\% confidence intervals) for the BT model when considering \BALToRc{}. 
While the target coverage $c$ varies, \BALToRc{} maintains stable the proportions of the classes in $\mcY = \{-1,0,1\}$.}
\label{fig:SRAres}
\end{figure*}

\begin{figure*}[t]
\centering
\begin{subfigure}[t]{0.24\textwidth}
    \includegraphics[scale=0.14]{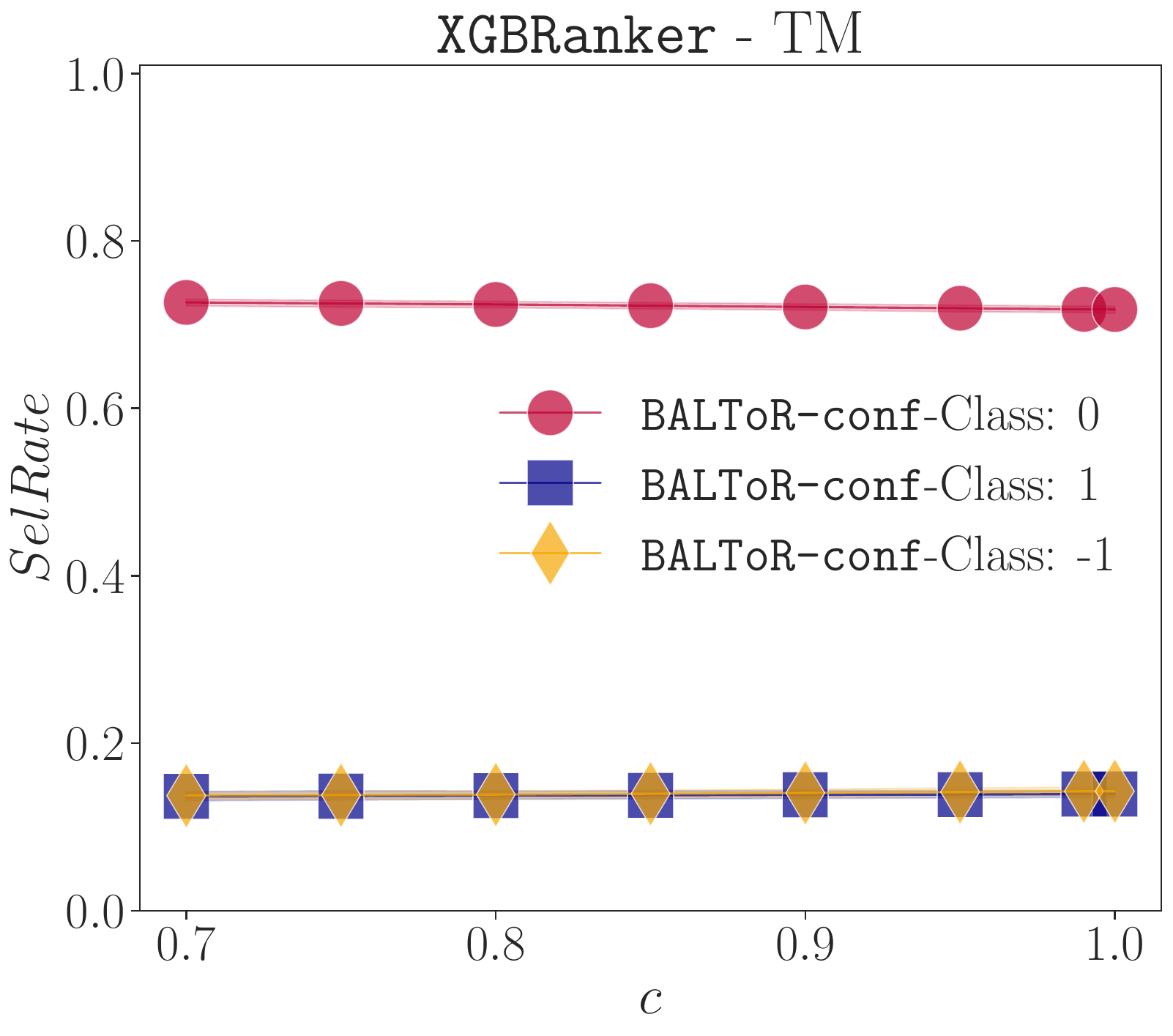}
    \caption{\texttt{MQ2007}}
    \label{fig:S1res_mq2007_TM}
\end{subfigure}
\hfill
\begin{subfigure}[t]{0.24\textwidth}
        \centering
    \includegraphics[scale=0.14]{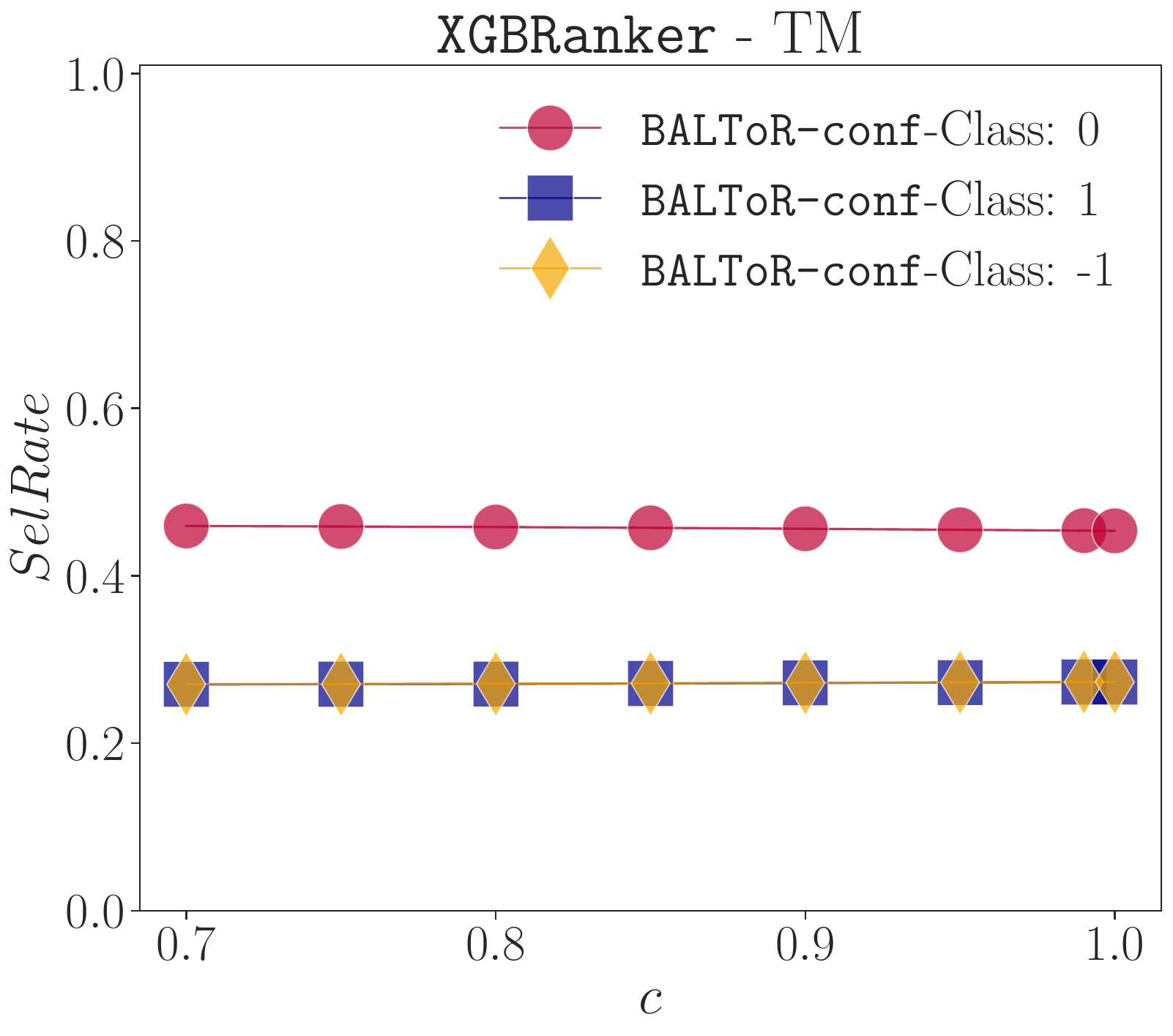}
    \caption{\texttt{Web-30k}}
    \label{fig:S1res_mslr_TM}
\end{subfigure}
\hfill
\begin{subfigure}[t]{0.24\textwidth}
        \centering
    \includegraphics[scale=0.14]{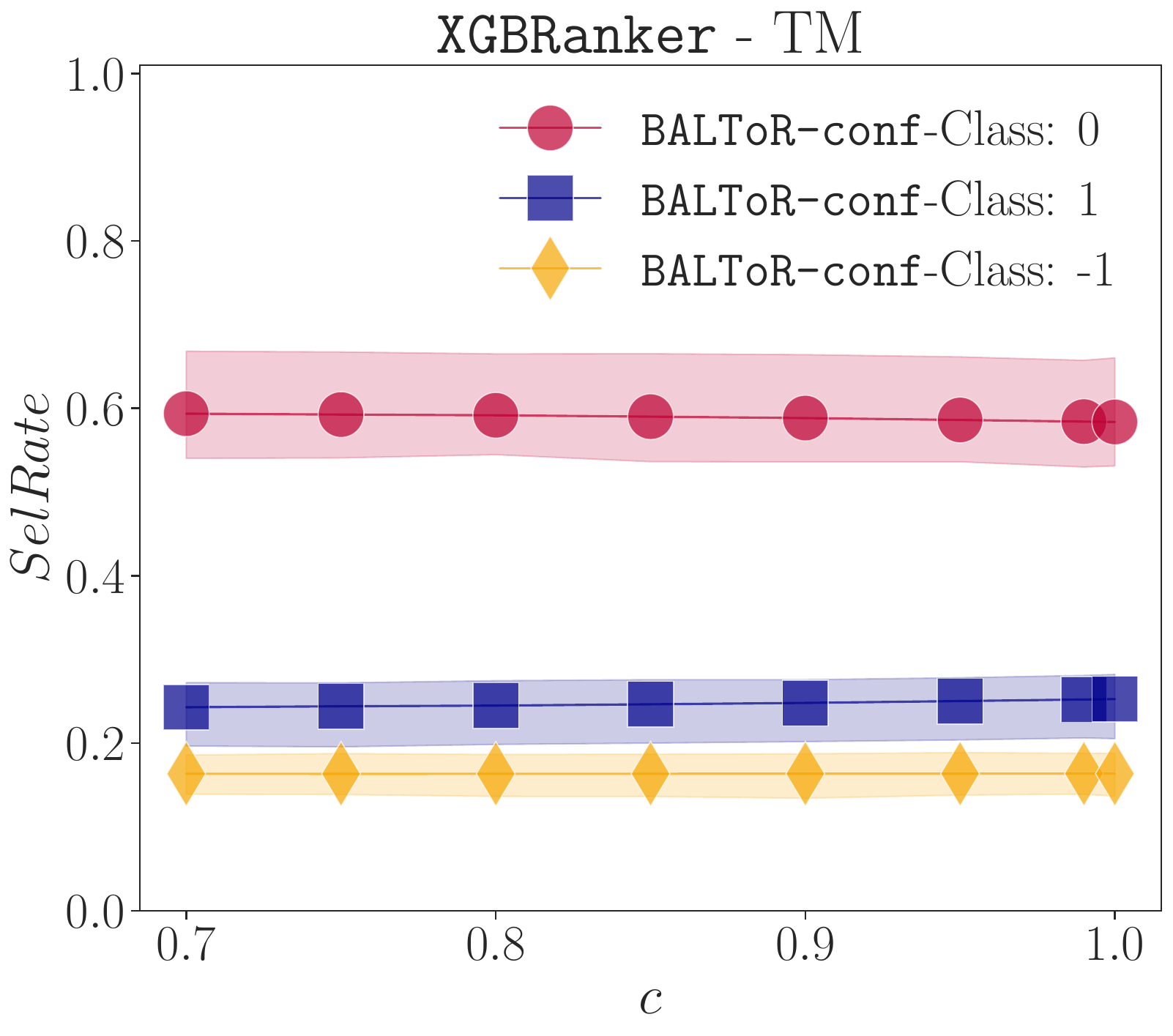}
    \caption{\texttt{OHSUMED}}
    \label{fig:S1res_ohsumed_TM}
\end{subfigure}
\hfill
\begin{subfigure}[t]{0.24\textwidth}
        \centering
    \includegraphics[scale=0.14]{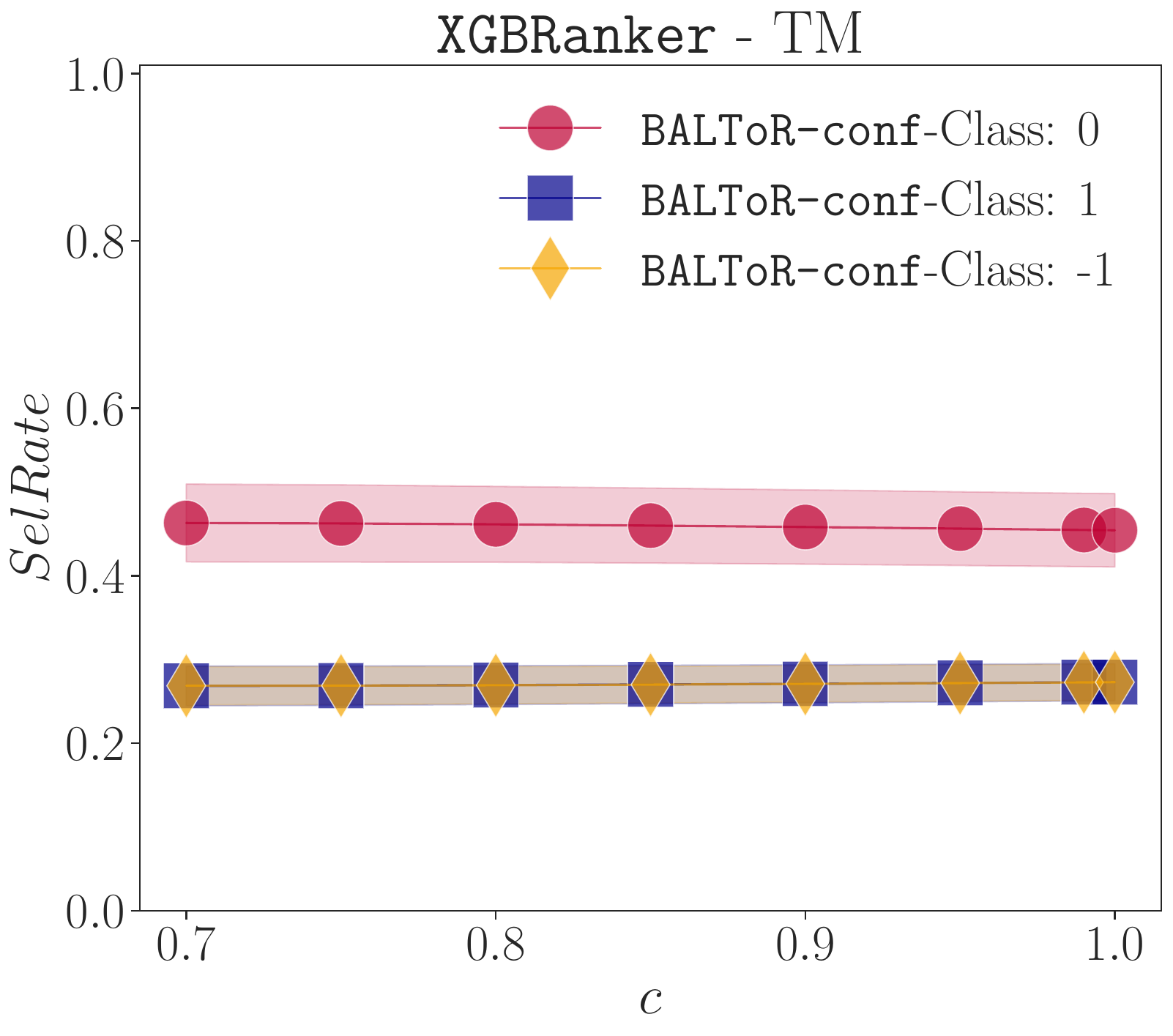}
    \caption{\texttt{Yahoo}}
    \label{fig:S1res_yahoo_TM}
\end{subfigure}
\caption{Class distribution $SelRate$ on the selected pairs (mean and 95\% confidence intervals) for the TM model when considering \BALToRc{}. 
While the target coverage $c$ varies, \BALToRc{} maintains stable the proportions of the classes in $\mcY = \{-1,0,1\}$.}
\label{fig:SRAres_TM}
\end{figure*}

In this section, we empirically assess the effectiveness of \BALToR{} using four popular and publicly available datasets for the learning-to-rank task. Our method aims to identify and abstain from pairs that are truly uncertain. By abstaining from a larger number of uncertain pairs, we expect improved performance on the remaining pairs. Additionally, we aim for the actual coverage to closely match the target coverage and for the rejections to maintain the overall distribution of classes in the pairwise comparisons.

More in details, we address the following questions:
\begin{itemize}
\item [\textbf{Q1:}] Which of the proposed approaches to estimate the conditional risk improves the most the accuracy on the pairwise ordering relationship?


    \item [\textbf{Q2:}] Is the coverage constraint satisfied at test time?
    \item [\textbf{Q3:}] How is rejection distributed across classes, i.e., is \BALToR{} (un)evenly rejecting pairs in which the first items is ranked higher than the second, or viceversa, or when there is a tie?
\end{itemize}
\subsection{Experimental Settings}\label{sec:subs_settings}
\paragraph{Datasets}~We consider four popular datasets from the Learning-to-Rank literature:
\texttt{Web-30k}, \texttt{OHSUMED}, \texttt{MQ2007} and \texttt{Yahoo} . 

\texttt{Web-30k} \cite{DBLP:journals/corr/QinL13} contains more than 30,000  queries, with each query having 125 assessed items/documents on average. Each query-document pair is encoded with 136 features. Furthermore,  relevance judgments are obtained from a retired labeling set of a commercial web search engine (Microsoft Bing), with values ranging from 0 (irrelevant) to 4 (perfectly relevant). 
\texttt{OHSUMED} \citep{hersh1994ohsumed} is a MEDLINE database subset containing 106 queries with 16,140 query-documents pairs. Each query-document pair has a 25-dimensional feature vector that contains popular information-retrieval features, such as tf-idf and BM25 score. There are three levels of relevance judgments: not relevant, possibly relevant and definitely relevant. 
\texttt{MQ2007} \citep{DBLP:journals/corr/QinL13} is a query set from the Million Query track of TREC 2007 and consists of 1,700 queries. The query-document pairs are represented with 46 features including scores such as BM25, PageRank, and HITS. The relevance labels are on a three-grade scale from 0 (irrelevant) to 2 (very relevant).
\texttt{Yahoo} \citep{DBLP:journals/jmlr/ChapelleC11} consists of features extracted from (query,url) pairs with relevance judgments. The training dataset contains $\approx19.5k$ queries and $\approx473k$ urls. Each url is given a relevance judgment with respect to the query. There are 5 levels of relevance from 0 (least relevant) to 4 (most relevant).

\paragraph{\BALToR\  variations and baselines}
We consider the default implementation of LambdaMART \citep{burges2010ranknet} from the popular libraries XGBoost \citep{DBLP:conf/kdd/ChenG16}, CatBoost \citep{prokhorenkova2018catboost} and LightGBM \citep{ke2017lightgbm}, as these are still state-of-the-art methods in learning to rank~\citep{DBLP:conf/iclr/0002YZTPWBN21}. We report the results for XGBoost in the main body of the paper. Since the results for CatBoost and LightGBM are very similar and consistent with those of XGBoost, we report them in the Supplementary Material.
To obtain probabilities for \cref{alg:est0-1}, we consider 
the Bradley-Terry (\BT) approach and the Thurstone-Mosteller (\TM) approach.
We consider the following approaches: $(i)$ a method using $1-\hat{p}_{max}$ (one minus the confidence) as the conditional risk estimator (\BALToRc{}); $(ii)$ a selection function that rejects pairs of instances thresholding the entropy of the predicted probabilities  (\BALToRe{}) rather than the conditional risk (the higher the entropy, the more uncertain the ranker); $(iii)$ a selection function based on the margin (\BALToRm). Moreover, since no methods for Bounded Abstention given a ranker are present in the literature, we consider also $(iv)$ a \textbf{random abstainer}, i.e., a selection function that selects a fraction $c$ of pairs of instances uniformly at random.

\paragraph{Metrics}
Concerning \textbf{Q1}, we evaluate the effectiveness of our approach in correctly comparing pairs of instances, by computing the accuracy $Acc$ 
(fraction of correctly ranked pairs) on the selected pairs at target coverage $c$. 
For \textbf{Q2}, we evaluate the ability to meet the coverage constraint by calculating the actual coverage $Cov = \frac{1}{\mid \mcD_{test}\mid}\sum \sum_{\mbx,\mbx'\in\mcD_{test}}g(\mbx,\mbx')$ at target coverage $c$.
Concerning \textbf{Q3}, we compute the fraction $SelRate$ of each class ($y \in \mcY = \{-1,0,1\}$) in the selected pairs of instances at target coverage 
$c$.

\paragraph{Setup}

For each experimental dataset, we train rankers on the training set. Then, for each target coverage $c\in\{.99, .95, .90, .85, .80, .75, .70\}$, we calibrate $\hat{\beta}_c$ on the calibration set.
Then, we compute the metrics of interest on the test set.
Each dataset provide five default folds (except \texttt{Yahoo} which contains two default folders) and in each fold the training, calibration (validation), and test splits are already predefined. 
Hence, we repeat the above procedure for each of the default folds provided in each dataset. 

We use default parameters for the learning-to-rank methods and we choose the tie probability parameter~$\theta = 2 \cdot n_{\text{pairs}} / n_{\text{no ties}} - 1$, where~$n_{\text{no ties}}$ is the number of pairs without ties and $n_{\text{pairs}}$ is the total number of pairs, as done in~\citep{rao1967ties}.

\paragraph{Reproducibility}
All experiments were run on a 64-core machine with AMD EPYC 7313 16-Core Processor, RAM 1TB, OS Ubuntu 22.04.5 LTS. The code can be retrieved at \textcolor{blue}{\url{https://github.com/Ambress92/Bounded-Abstention-LTR}.}

\subsection{Experimental Results}

\paragraph{Q1: \BALToRm{} and \BALToRc{} allow for improvements in performance.} Consider \cref{fig:AccResBT,fig:AccResTM}, which report the accuracy at target coverage $c$. Notably, the accuracy of the random abstainer remains flat across all datasets and probability models. Conversely, \BALToRm{} and \BALToRc{} show an increasing accuracy when decreasing $c$'s, confirming the effectiveness in correctly detecting when to abstain. 

Overall, we observe that when using the BT model, \BALToRm{} is the best performer on all four datasets, with significant results on \texttt{MSLR-30k}, for all $c\leq .80$. The second best performer is \BALToRc{}, which achieves comparable performance to \BALToRm{} on \texttt{MQ2007}, \texttt{OHSUMED} and \texttt{Yahoo}.
Interestingly, we see that \BALToRe{} performs poorly on all the datasets, with worse performance than random. 
Conversely, when looking at TM, \BALToRc{} and \BALToRm{} have closer performance, with \BALToRc{} achieving slightly better (but not significant) improvements over \BALToRm{}. Interestingly, also for \BALToRe{}, we see improvements in all datasets except for \texttt{MSLR-30k}.

A reason for the \BALToRe{} erratic behaviour is the fact that the entropy density function might differ from the risk density function. This depends on how smoothed the predicted probabilities are, hence not always capturing when to abstain correctly. This is especially true for the BT model, which tends to produce flatter probabilities compared to the TM model. We showcase an example of this erratic behaviour of the entropy baseline in \cref{fig:example_entropy}, which provides the plots for the estimated density functions over \texttt{MQ2007} Fold 1 calibration set for both entropy and \BALToR{}. On the one hand, we can observe that the shapes of the density functions differ when using the BT model, thus the entropy-baseline fails to correctly detect pairs to abstain on. On the other hand, when using TM model, the entropy's distribution is similar to \BALToR{}'s, thus achieving similar results.
We argue this pattern occurs as the BT model tends to provide predicted probabilities that are less peaked, hence increasing the overall entropy.

These results suggest that \BALToRc{} and \BALToRm{} are the best approaches to perform abstention on pairwise ranking comparisons.

\paragraph{Q2: coverage constraint is almost always satisfied.}
A key requirement for abstention strategies is that selection functions must respect the coverage constraint of our problem formulation (\Cref{eq:l2rank_abst}).
In \cref{fig:COVres} we report the values for the \texttt{XGBRanker} when using BT, while \cref{fig:COVresTM} depicts the results for the TM model. We provide results for other rankers in the Appendix.

In all cases, $Cov$ is very close to the target coverage $c$. For \texttt{MQ2007}, \texttt{Web-30k} and \texttt{Yahoo}, the actual coverage distance from $c$ is less than $\approx.001$ regardless of the probability model and uncertainty measure considered.
The average distance is slightly larger for \texttt{OHSUMED}, with the largest difference of $\approx .002$ at $c=.75$ and $c=.70$.

Overall,  results confirm the effectiveness of our approach in satisfying the coverage constraint.

\paragraph{Q3: all methods exhibit stable acceptance ratios.}

A critical aspect that is often underestimated in the literature is the ability of abstaining systems to keep stable distribution of classes $y \in \mcY = \{-1,0,1\}$ in the selected (pairs of) instances, where $-1$ means that the second instance has a higher relevance than the first, $1$ indicates that the first is preferred and $0$ a tie). 
Indeed, abstention strategies that concentrate on one class might introduce forms of biases in the decision-making process~\citep{Benchmark2024}. 

\cref{fig:SRAres} and \cref{fig:SRAres_TM} report the $SelRate$ at different coverages $c$ when considering \BALToRc{} with BT and TM models respectively. For the sake of space, we provide the results for the other methods and other rankers in \cref{Appendix:additionalresults}.

We can see that all methods show stable $SelRate$ when varying $c$.
On \texttt{MQ2007}, when considering the BT model, the $SelRate$ for the $0-$class pairs ranges between $\approx .718\pm .005$ and $\approx.725\pm.005$, the $-1$-class $SelRate$ is between $\approx.143\pm.007$ and $\approx .138  \pm.007$ and the $-1$-class $SelRate$ is stable around $\approx .139\pm.007$ and $\approx .136\pm .007$. 
On \texttt{Web-30k}, for the BT model, the $SelRate$ remain stable around $\approx.272\pm.002$ both for class $-1$ and $1$, while the average $SelRate$ for class $0$ is $\approx.456\pm.004$. Similar results also hold for TM.
On \texttt{OHSUMED}, the $SelRate$ remains almost flat
for both the BT and the TM models: the former's $SelRate$ is $\approx .163\pm.033$ for class $-1$, $\approx.249\pm.002$ for class $1$, and $.586\pm.082$ for class $0$;
while the latter achieves a $SelRate$ of $\approx.248\pm.051$ for class $1$, $\approx .589 \pm .082$ for class $0$ and $\approx .163\pm.033$ for class $-1$.
On \texttt{Yahoo}, the $SelRate$ is flat for BT: for class $-1$ $SelRate\approx.277\pm.036$, class $1$ $SelRate\approx.277\pm.037$ and class $0$ $SelRate\approx.446 \pm .074$. Similar results are also achieved by the TM model, with similar levels of performance.

Overall, these results confirm that the rejections are evenly distributed across the classes and show no biases in rejecting pairs of instances belonging to specific classes.

\section{Conclusions}
\label{sec:conclusion}

Abstention methods have proven effective in mitigating the effects of uncertainty in ML models. In this paper, we introduced the bounded-abstention framework in the context of pairwise learning-to-rank. We provided a theoretical characterization of the optimal selection strategy and proposed a model-agnostic algorithm to implement it. Our experimental results demonstrated the effectiveness of the approach, showing improvements in accuracy and ranking performance, while satisfying coverage constraints and maintaining stable selection rates across different classes.

\paragraph{Limitations} 
Our implementation relies on a given ranker provided by a learning-to-rank algorithm. In practice, obtaining a ranker with strong performances can be challenging. In turn, this could negatively affect the quality of the selection strategy, with limited improvements when abstention occurs, as known in the classification task \citep{DBLP:journals/jmlr/FrancPV23,Benchmark2024}. While we considered the problem of optimizing the selection function $g$, with the ranker $f$ given, an end-to-end approach is worth considering, where both $f$ and $g$ are optimized together.

\paragraph{Future Works} We envision several future research directions. First, we aim at studying how to build optimal abstention strategy for other ranking tasks, such as~listwise ranking.  
Second, our work focuses on ambiguity rejection. We intend to explore approaches targeting novelty rejection in the learning-to-rank setting by abstaining on pairs that are out-of-distribution.
Finally, we implicitly assume that humans can correctly break ties when asked to. Such an assumption is challenged in the learning-to-defer framework~\citep{DBLP:conf/nips/MadrasPZ18}, where humans can also make mistakes. Considering strategies for learning to defer in the context of ranking is an unexplored area.

\paragraph{Broader Impact}
This paper focuses on abstention mechanisms in learning to rank. The goal is to allow a ML system to avoid low-confidence decision and  inform a human decision maker, who can then intervene directly or help the system to make a more informed decision. While in this paper we mainly focus on the abstention framework,  eventual biases might especially arise from the interplay between algorithms and human decision making. We do not envision any direct ethical concern arising from our work, however a special attention might be necessary when abstention mechanisms are used in high-stakes domains. 

\begin{acks}
    The research of Andrea Pugnana and Salvatore Ruggieri was supported by the Horizon Europe Programme, grant \#101120763-TANGO.
Funded by the European Union. Views and opinions expressed are however those of the author(s) only and do not necessarily reflect those of the European Union or the European Health and Digital Executive Agency (HaDEA). Neither the European Union nor the granting authority can be held responsible for them.
\end{acks}

\balance
\bibliographystyle{ACM-Reference-Format}
\bibliography{biblio}

\clearpage
\appendix
\newpage
\section{Proofs}
\label{Appendix:proofs}
\subsection{Proof of Theorem 3.1}
\label{prf:thm_3_1}
\begin{proof}

First, consider that:
condition p1 implies that
\[g^*(\mbx,\mbx') = 1 \quad \forall (\mbx,\mbx')\in \riskLowerBeta\] almost-surely. As $p(\mbx,\mbx')\geq 0$ and since $g(\mbx,\mbx')\in[0,1]$ it must always be one on all the points with positive probability mass in $\riskLowerBeta$ to make p1 hold.

Similarly, condition p3 implies that
\[g^*(\mbx,\mbx') = 0 \quad \forall (\mbx,\mbx')\in \riskGreaterBeta\] almost-surely. As $p(\mbx,\mbx')\geq 0$ and since $g(\mbx,\mbx')\in[0,1]$ it must always be zero on all the points with positive probability mass in $\riskLowerBeta$ to make p3 hold.

Now, let us consider $R_l(f,g,y)$.
This can be decomposed as:
\[R_l(f,g,y) = \frac{\mathbb{E}_{(\mbx, \mbx',y)}[l(f(\mbx), f(\mbx'),y)g(\mbx,\mbx')]}{\mathbb{E}_{(\mbx,\mbx')}[g(\mbx, \mbx')]},
\]
where the numerator is
\[
\begin{split}
\mathbb{E}_{(\mbx, \mbx',y)}&[l(f(\mbx), f(\mbx'),y)g(\mbx,\mbx')] \\
& =   {\iint}_{\mcX\times\mcX}p(\mbx,\mbx')g(\mbx,\mbx')\biggl(\\ &\qquad\int_{\mcY\mid\mcX\times\mcX}p(y\mid\mbx,\mbx')l\left(f(\mbx'),f(\mbx'),y\right)dy\biggr) d\mbx d\mbx'\\
&={\iint}_{\mcX\times\mcX}p(\mbx,\mbx')g(\mbx,\mbx')r(\mbx,\mbx') d\mbx d\mbx',
\end{split}\]
and the denominator is
\[\mathbb{E}_{(\mbx,\mbx')}[g(\mbx, \mbx')] = \iint_{\mcX\times\mcX}p(\mbx,\mbx')g(\mbx,\mbx')d\mbx d\mbx' \;. \]

If conditions \textit{(p1), (p2)} and \textit{(p3)} are met, then for every $f\in\mathcal{F}$, $g^*:\mcX\times\mcX\to [0,1]$ satisfies:
\begin{equation}
\begin{split}  
    &R(f,g^*,y) = \\
    & \beta + \frac{1}{c}\iint_{\riskLowerBeta}p(\mbx,\mbx')d\mbx d\mbx'-\frac{\beta}{c}\iint_{\riskLowerBeta}p(\mbx,\mbx')d\mbx d\mbx',
    \end{split}
    \end{equation}
as it holds that
\[
\begin{split}
&{\iint}_{\mcX\times\mcX}p(\mbx,\mbx')g^*(\mbx,\mbx')r(\mbx,\mbx') d\mbx d\mbx' =\\
&= {\iint}_{\riskLowerBeta}p(\mbx,\mbx')g^*(\mbx,\mbx')r(\mbx,\mbx') d\mbx d\mbx' +\\
&\qquad+
{\iint}_{\riskEqualBeta}p(\mbx,\mbx')g^*(\mbx,\mbx')r(\mbx,\mbx') d\mbx d\mbx' +\\
&\qquad+{\iint}_{\riskGreaterBeta}p(\mbx,\mbx')g^*(\mbx,\mbx')r(\mbx,\mbx') d\mbx d\mbx' =\\
&={\iint}_{\riskLowerBeta}p(\mbx,\mbx')g^*(\mbx,\mbx')r(\mbx,\mbx') d\mbx d\mbx' +\\
&\qquad +\beta
{\iint}_{\riskEqualBeta}p(\mbx,\mbx')g^*(\mbx,\mbx')d\mbx d\mbx'=\\
&={\iint}_{\riskLowerBeta}p(\mbx,\mbx')g^*(\mbx,\mbx')r(\mbx,\mbx') d\mbx d\mbx' +\\
&\qquad + \beta[ c -
{\iint}_{\riskLowerBeta}p(\mbx,\mbx')d\mbx d\mbx']=\\
&= \beta c +
{\iint}_{\riskLowerBeta}p(\mbx,\mbx')(g^*(\mbx,\mbx')r(\mbx,\mbx') -\beta)d\mbx d\mbx'= \\
&=\beta c + {\iint}_{\riskLowerBeta}p(\mbx,\mbx')(r(\mbx,\mbx') -\beta)d\mbx d\mbx' ,
\end{split}
\]
where the last equality comes from the fact that $g^*(\mbx,\mbx')$ is one on all the points in $\riskLowerBeta$.
Concerning the denominator, we can see that it is equal to:
\[
\begin{split}
\mathbb{E}&_{(\mbx,\mbx')}[g^*(\mbx, \mbx')] =\\
&=
\iint_{\mcX\times\mcX_{r(\mbx, \mbx')<\beta}} p(\mbx,\mbx')d\mbx d\mbx' +\\
& \qquad + c -\iint_{\mcX\times\mcX_{r(\mbx, \mbx')<\beta}} p(\mbx,\mbx')d\mbx d\mbx' = c
\end{split}
\]
Hence, $g^*$ is feasible.
To prove the theorem, we show that for each $g$ that is feasible (i.e.,~$g$ is such that $\phi(g)\geq c$)
and does not satisfy one of the conditions, we can always define another $g{'}$ with a lower risk.
We first show that if we consider $g$ such that p1 is violated, but $p2$ and $p3$ hold, then $g$ is not optimal.
Then, we show that if $p1$ and $p3$ holds but $p2$ is violated, $g$ is not optimal.
Then, we show that if $p3$ does not hold, we also have a violation of $p1$.

\textbf{Case 1: violation of p1.}
Let us assume that p2 and p3 hold.
If p1 is violated, it means that:

\begin{equation}
    \iint_{\riskLowerBeta} p(\mbx,\mbx')g(\mbx,\mbx') d\mbx d\mbx' < \iint_{\riskLowerBeta}p(\mbx,\mbx') d\mbx d\mbx'
\end{equation}

This implies that $\exists \tilde{X} \subseteq \mcX\times\mcX \quad s.t.$
\begin{equation}
\label{eq:Viol1}
    \forall (\mbx,\mbx')\in \tilde{X}:\,r(\mbx,\mbx')\geq \beta
\end{equation}
Alternatively, we can think of $\tilde{X}$ as the set of points for which it holds that:
$(\mbx, \mbx') \in \mcX\times \mcX:  0 < g(\mbx, \mbx') < 1$

and
\begin{equation}
    \label{eq:Viol2}
    \begin{split}
            \iint_{\tilde{X}}&p(\mbx,\mbx')g(\mbx,\mbx') d\mbx d\mbx' = \\
    &=\iint_{\riskLowerBeta}p(\mbx,\mbx')d\mbx d\mbx' -\\
    &\qquad - \iint_{\riskLowerBeta}p(\mbx,\mbx')g(\mbx,\mbx')d\mbx d\mbx' > 0
    \end{split}
\end{equation}

Now, let us consider the following $g'$:
\begin{equation}
g'(\mbx,\mbx') =
    \begin{cases}
        1 \quad \text{if} \quad r(\mbx,\mbx')<\beta\\
        0 \quad \text{if} \quad (\mbx,\mbx') \in \tilde{X}\\
        g(\mbx,\mbx') \quad \text{otherwise}
    \end{cases}
\end{equation}

Now notice that $g'$ is feasible, as
\[
\begin{split}
    \phi(g') &= \iint_{\mcX\times\mcX}p(\mbx,\mbx')g'(\mbx,\mbx') d\mbx d\mbx' = \\
    &=\iint_{\riskLowerBeta}p(\mbx,\mbx')d\mbx d\mbx' +\\
&\qquad + \iint_{\riskEqualBeta}g(\mbx,\mbx')p(\mbx,\mbx') = \phi(g)
\end{split}
\]

Let us now compute $\phi(g)\left(R(f,g,y) - R(f,g',y)\right)$:

\[\begin{split}
    \phi(g)&\left(R(f,g,y) - R(f,g',y)\right)= \\
    &=\iint_{\mcX\times\mcX}g(\mbx,\mbx')p(\mbx,\mbx')r(\mbx,\mbx')d\mbx d\mbx'-\\ 
    &\qquad -\iint g'(\mbx,\mbx) p(\mbx,\mbx') r(\mbx,\mbx')d\mbx d\mbx'=\\
    &= \iint_{\mcX\times\mcX}g(\mbx,\mbx')p(\mbx,\mbx')r(\mbx,\mbx')d\mbx d\mbx' -\\ 
    &\qquad -\iint_{\riskLowerBeta}p(\mbx,\mbx')r(\mbx,\mbx')d\mbx d\mbx' - \\ 
    &\qquad -\iint_{\riskGreaterEqualBeta}g(\mbx,\mbx')p(\mbx,\mbx')r(\mbx,\mbx')d\mbx d\mbx'=\\
    &= \iint_{\riskLowerBeta}g(\mbx,\mbx')p(\mbx,\mbx')r(\mbx,\mbx')d\mbx d\mbx' +\\ 
    &\qquad + \iint_{\tilde{X}}g(\mbx,\mbx')p(\mbx,\mbx')r(\mbx,\mbx')d\mbx d\mbx' -\\ 
    &\qquad -  \iint_{\riskLowerBeta}p(\mbx,\mbx')r(\mbx,\mbx')d\mbx d\mbx' =\\
    &=
    \iint_{\riskLowerBeta}\left(
    g(\mbx,\mbx')-1
    \right)p(\mbx,\mbx')r(\mbx,\mbx')d\mbx d\mbx'+ \\ 
    &\qquad + \iint_{\tilde{X}}g(\mbx,\mbx')p(\mbx,\mbx')r(\mbx,\mbx')
    \end{split}\]

Now, we can re-write the last equation as follows, by noticing that Eq. \ref{eq:Viol1} and Eq. \ref{eq:Viol2} imply that:
\[
\begin{split}
&\iint_{\riskLowerBeta}\left(
    g(\mbx,\mbx')-1
    \right)p(\mbx,\mbx')r(\mbx,\mbx')d\mbx d\mbx'+\\ 
    &\qquad +
    \iint_{\tilde{X}}g(\mbx,\mbx')p(\mbx,\mbx')r(\mbx,\mbx') \geq \\
    & \geq
    \iint_{\riskLowerBeta}\left(
    g(\mbx,\mbx')-1
\right)p(\mbx,\mbx')r(\mbx,\mbx')d\mbx d\mbx' +\\ 
    &\qquad + \beta
    \iint_{\riskLowerBeta}\left(1-g(\mbx,\mbx')\right)p(\mbx,\mbx') d\mbx d\mbx' =\\
    & = \iint_{\riskLowerBeta}p(\mbx,\mbx')\left(1-g(\mbx,\mbx')\right)\left(\beta - r(\mbx,\mbx')\right) d\mbx d\mbx'>0
\end{split}
\]
Hence, if $p1$ is not holding, $g(\mbx,\mbx')$ is not optimal.

\textbf{Case 2: violation p2}
Let us assume that $p1$ is satisfied, but $p2$ is violated, i.e.,
\[
\iint_{\riskEqualBeta}g(\mbx,\mbx')p(\mbx,\mbx')d\mbx d\mbx' < c - \iint_{\riskLowerBeta}p(\mbx,\mbx')d\mbx d\mbx' > 0\]

In this case, we can define a $g':\mcX\times\mcX \to [0,1]$ such that
\begin{equation}
g'(\mbx,\mbx') = \begin{cases}
  g(\mbx,\mbx') \quad \text{if} \quad r(\mbx,\mbx') < \beta \\
  0 \quad \text{if} \quad r(\mbx,\mbx')>\beta
\end{cases}
\end{equation}
and it satisfies the following condition:
\begin{equation}
\begin{split}
        \iint_{\riskEqualBeta}& g'(\mbx,\mbx)p(\mbx,\mbx')d\mbx d\mbx' = \\
        &=\iint_{\riskEqualBeta}g(\mbx,\mbx')p(\mbx,\mbx')d\mbx d\mbx' +\\ 
    &\qquad +
    \iint_{\riskGreaterBeta}g(\mbx,\mbx')p(\mbx,\mbx') d\mbx d\mbx'
\end{split}
\end{equation}

Now, we can see that $\phi(g)=\phi(g')$ as:
\[
\begin{split}
    \phi(g')& = \iint_{\mcX\times\mcX} g'(\mbx,\mbx) p(\mbx,\mbx) d\mbx d\mbx' = \\
    &= \iint_{\riskLowerBeta}g'(\mbx,\mbx')p(\mbx,\mbx')d\mbx  d\mbx' + \\ 
    &\qquad +
\iint_{\riskEqualBeta}g'(\mbx,\mbx')p(\mbx,\mbx')d\mbx  d\mbx' = \\
&=\iint_{\riskLowerBeta}g(\mbx,\mbx')p(\mbx,\mbx')d\mbx  d\mbx' +\\ 
    &\qquad +
\iint_{\riskEqualBeta}g(\mbx,\mbx')p(\mbx,\mbx')d\mbx d\mbx' +\\ 
    &\qquad  +\iint_{\riskGreaterBeta}g(\mbx,\mbx')p(\mbx,\mbx') d\mbx d\mbx' = \phi(g)
\end{split}
\]

Hence, we can consider the following:
\[
\begin{split}
\phi(g)&\left(R(f,g,y) - R(f,g',y)\right) = \\
&=
\beta \iint_{\riskEqualBeta}g(\mbx,\mbx') p(\mbx,\mbx') d\mbx d\mbx' +\\ 
    &\qquad + \iint_{\riskGreaterBeta}
g(\mbx,\mbx') p(\mbx,\mbx') r(\mbx, \mbx') d\mbx d\mbx'
-\\ 
    &\qquad -
\beta \iint_{\riskEqualBeta}g'(\mbx,\mbx') p(\mbx,\mbx') d\mbx d\mbx' =\\
&=\iint_{\riskGreaterBeta}g(\mbx,\mbx')p(\mbx,\mbx')r(\mbx,\mbx')d\mbx d\mbx' - \\ 
    &\qquad - \beta \iint_{\riskGreaterBeta}g(\mbx,\mbx') p(\mbx,\mbx') d\mbx d\mbx' >\\
&> \beta \iint_{\riskGreaterBeta}g(\mbx,\mbx')p(\mbx,\mbx')d\mbx d\mbx' - \\ 
    &\qquad - \beta \iint_{\riskGreaterBeta}g(\mbx,\mbx') p(\mbx,\mbx') d\mbx d\mbx'=0
\end{split}
\]
Hence, $g$ is not optimal.

\textbf{Case 3. Violation of $p2$ or $p3$.}
Let us consider the case when $g$ is such that $\phi(g)>c$, i.e., either one of the following conditions hold:
\begin{equation}
    \iint_{\riskEqualBeta} g(\mbx,\mbx')p(\mbx,\mbx')d\mbx d\mbx' > c - \iint_{\riskLowerBeta}p(\mbx,\mbx')d \mbx d \mbx'
\end{equation}
\begin{equation}
    \iint_{\riskGreaterBeta}g(\mbx,\mbx')p(\mbx,\mbx')d \mbx d \mbx' > 0
\end{equation}

We notice that for every $\alpha\in\mathbb{R}^+$, it holds that $R(f,g,y)=R(f,\alpha \cdot g, y)$.
Now we consider $g':\mcX\times\mcX\to [0,1]$ such that $g' = \frac{c}{\phi(g)}g$.
It is straightforward to notice that $\phi(g') = c$ and the following holds:
\begin{equation}
\begin{split}
\iint_{\riskLowerBeta}&g'(\mbx,\mbx')p(\mbx,\mbx')d\mbx d\mbx' = \\ &=\frac{c}{\phi(g)}\iint_{\riskLowerBeta}g(\mbx,\mbx')p(\mbx,\mbx')d\mbx d\mbx' <\\
&< \iint_{\riskLowerBeta}p(\mbx,\mbx')d\mbx d\mbx'
    \end{split}
\end{equation}

This implies that $g'$ violates condition $p1$ and is therefore not optimal.
Consequently, since $R(f,g',y)=R(f,g, y)$, $g$ is also not optimal.
\end{proof}

\subsection{Proof of Theorem 3.2}
\label{prf:thm_3_2}
\begin{proof}
    Clearly, $g^*$ satisfies $p1$ and $p3$.
    If $\iint_{\riskEqualBeta}p(\mbx,\mbx')d\mbx d \mbx'=0$, then condition $p2$ is met as $\iint_{\riskLowerBeta}p(\mbx,\mbx')d\mbx d \mbx'=c$.
    Otherwise, by the definition of $g^*$ it holds that:
\[
\begin{split}
    \iint_{\riskEqualBeta}&\left(
c -\frac{\iint_{\riskLowerBeta}p(\mbz,\mbz')d\mbz d\mbz'}{\iint_{\riskEqualBeta}p(\mbz,\mbz')d\mbz d\mbz'}
\right)
p(\mbx,\mbx')d\mbx d\mbx' = \\ 
& = 
c-{\iint_{\riskLowerBeta}p(\mbx,\mbx')d\mbx d\mbx'}
\end{split}
\]
which means condition $p2$ is met. Hence, $g^*$ is optimal.
\end{proof}

\subsection{Proof of Proposition 3.3}
\label{prf:pro_3_3}
\begin{proof}

    Let us consider the 0-1 loss, hence the conditional risk is:
    \[
    \begin{split}
        r&_{0-1}(\mbx,\mbx') = \\ & = p(1\mid\mbx,\mbx')\mathbb{I}_{\hat{y}={-1,0}} +
    p(0\mid\mbx,\mbx')\mathbb{I}_{\hat{y}={-1,1}}+
    p(-1\mid\mbx,\mbx')\mathbb{I}_{\hat{y}={0,1}},
    \end{split}
    \]
    where $$\hat{y} = \argmax_{y\in\{-1,0,1\}} \{p(-1\mid\mbx,\mbx'), p(0\mid\mbx,\mbx'), p(1\mid\mbx,\mbx')\}$$ is the predicted label by the pairwise ranker.

    Now let us consider the case where $$p(-1\mid \mbx, \mbx') = \max_{y\in\{-1,0,1\}}\{p(-1\mid\mbx,\mbx'), p(0\mid\mbx,\mbx'), p(1\mid\mbx,\mbx')\}$$. 
    
    Then it holds that:
    \[
    \begin{split}
    r_{0-1}(\mbx,\mbx') =
p(0\mid\mbx,\mbx')+p(1\mid\mbx,\mbx') = 1- p(-1\mid \mbx, \mbx') = \\1-\max_{y\in\{-1,0,1\}}\{p(-1\mid\mbx,\mbx'), p(0\mid\mbx,\mbx'), p(1\mid\mbx,\mbx')\},
    \end{split}
    \]
    where the first equality comes from the fact that the prediction is $\hat{y}=-1$. By considering the remaining two cases, we can see that the result holds also when $$p(0\mid \mbx, \mbx') = \max_{y\in\{-1,0,1\}}\{p(-1\mid\mbx,\mbx'), p(0\mid\mbx,\mbx'), p(1\mid\mbx,\mbx')\}$$ and $$p(1\mid \mbx, \mbx') = \max_{y\in\{-1,0,1\}}\{p(-1\mid\mbx,\mbx'), p(0\mid\mbx,\mbx'), p(1\mid\mbx,\mbx')\},$$ concluding the proof.
\end{proof}

\section{Additional results}
\label{Appendix:additionalresults}

The additional results for $(i)$ $SelRate$ when using \BALToRe{} and \BALToRm{} based on the XGBoost ranker; $(ii)$ for all the metrics when using the CatBoost and LightGBM implementations of ranking can be found in the Supplementary Material at the following link \textcolor{blue}{\url{https://github.com/Ambress92/Bounded-Abstention-LTR}.}. Overall results are comparable, with a small difference among the different implementations.

\end{document}